\documentclass[10pt,twocolumn,letterpaper]{article}
\PassOptionsToPackage{table}{xcolor}

\usepackage[pagenumbers]{cvpr}

\usepackage{cuted} 

\usepackage{currfile} 

\usepackage{caption} 

\usepackage{multirow}
\newcounter{algorithm}
\newcounter{alglineno}
\newcounter{algdepth}
\newlength{\algindentsize}
\newcommand{\algcaption}[1]{%
  \refstepcounter{algorithm}%
  \par\noindent\textbf{Algorithm~\thealgorithm}~#1\par
  \vspace{0.22em}\hrule height 0.4pt\vspace{0.32em}%
}
\newenvironment{algorithm}[1][]{%
  \par\addvspace{0.45em}%
  \noindent\begin{minipage}{\linewidth}%
  \hrule height 0.7pt\vspace{0.28em}%
}{%
  \par\vspace{0.15em}\hrule height 0.7pt
  \end{minipage}%
  \par\addvspace{0.55em}%
}
\newenvironment{algorithmic}{%
  \setcounter{alglineno}{0}%
  \setcounter{algdepth}{0}%
  \normalsize\raggedright
  \ltopdlonglastfalse
  \setlength{\parindent}{0pt}%
  \setlength{\parskip}{0.22em}%
  \setlength{\rightskip}{0.4em}%
  \newcommand{\State}{%
    \par\noindent
    \refstepcounter{alglineno}%
    \makebox[1.55em][r]{\normalsize\arabic{alglineno}}%
    \hspace{0.45em}%
    \hspace*{\dimexpr\value{algdepth}\algindentsize\relax}%
    \ignorespaces
  }%
  \newcommand{\Require}{\State\textbf{Require:} }%
  \newcommand{\Ensure}{\State\textbf{Ensure:} }%
  \newcommand{\Return}{\State\textbf{return} }%
  \newcommand{\If}[1]{\State\textbf{if} ##1 \textbf{then}\addtocounter{algdepth}{1}}%
  \newcommand{\Else}{\addtocounter{algdepth}{-1}\State\textbf{else}\addtocounter{algdepth}{1}}%
  \newcommand{\EndIf}{\addtocounter{algdepth}{-1}\State\textbf{end if}}%
  \newcommand{\While}[1]{\State\textbf{while} ##1 \textbf{do}\addtocounter{algdepth}{1}}%
  \newcommand{\EndWhile}{\addtocounter{algdepth}{-1}\State\textbf{end while}}%
}{}

\newcommand{\venue}[1]{{\scriptsize\textcolor{gray}{(#1)}}}

\newcommand{\qwenlogo}{\raisebox{-0.32em}{\includegraphics[height=1.15em]{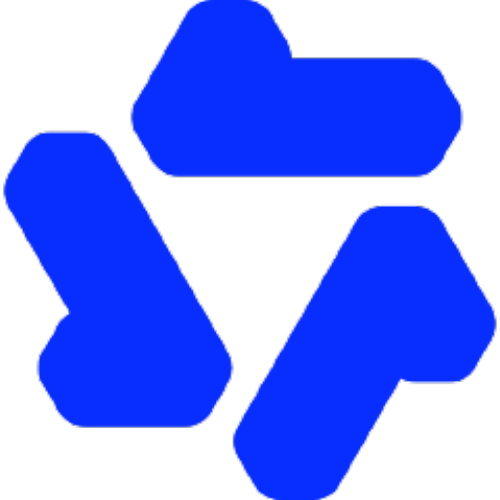}}\hspace{0.32em}}
\newcommand{\gemmalogo}{\raisebox{-0.32em}{\includegraphics[height=1.15em]{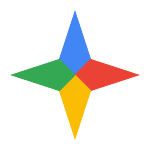}}\hspace{0.32em}}
\newcommand{\qwenhead}[1]{%
  \textit{#1} & \multicolumn{14}{c|}{\qwenlogo\textit{Qwen3.5-4B}}%
}
\newcommand{\gemmahead}[1]{%
  \textit{#1} & \multicolumn{14}{c|}{\gemmalogo\textit{Gemma~4 E4B-it}}%
}

\newcommand{\KL}{\mathrm{KL}}

\definecolor{cvprblue}{rgb}{0.21,0.49,0.74}
\usepackage{xr-hyper}
\usepackage[pagebackref,breaklinks,colorlinks,allcolors=cvprblue]{hyperref}
\AtBeginDocument{%
  \crefname{algorithm}{Alg.}{Algs.}%
  \Crefname{algorithm}{Alg.}{Algs.}%
  \Crefname{section}{Sec.}{Secs.}%
  \Crefname{table}{Tab.}{Tabs.}%
  \crefname{figure}{Fig.}{Figs.}%
  \Crefname{figure}{Fig.}{Figs.}%
  \crefname{equation}{Eq.}{Eqs.}%
  \Crefname{equation}{Eq.}{Eqs.}%
}

\usepackage{etoolbox}
\makeatletter
\newif\ifltopdlonglast
\ltopdlonglasttrue
\AtBeginDocument{%
  \pretocmd{\@sect}{\ltopdlonglastfalse}{}{}%
  \apptocmd{\@sect}{\ltopdlonglasttrue}{}{}%
  \pretocmd{\@ssect}{\ltopdlonglastfalse}{}{}%
  \apptocmd{\@ssect}{\ltopdlonglasttrue}{}{}%
}
\newif\ifltopdbib
\AtBeginEnvironment{thebibliography}{\global\ltopdbibtrue}
\apptocmd{\thebibliography}{%
  \frenchspacing
  \tolerance=1000
  \emergencystretch=3em
  \spaceskip=0pt
  \xspaceskip=0pt
}{}{}
\AtEndEnvironment{thebibliography}{%
  \global\ltopdbibfalse
  \global\tolerance=200
  \global\emergencystretch=2em
  \global\spaceskip=0pt
  \global\xspaceskip=0pt
}
\AtBeginDocument{%
  \pretocmd{\BR@backref}{\unskip\nobreak\ \mbox\bgroup}{}{}%
  \patchcmd{\BR@backref}{\endgroup\par}{\endgroup\egroup\par}{}{}%
}
\AddToHook{para/begin}{%
  \ifltopdlonglast
    \ifnum\gluestretchorder\rightskip=0
      \ifnum\gluestretchorder\leftskip=0
        \ifltopdbib
          \global\parfillskip=0pt plus 1fil\relax
          \global\emergencystretch=3em\relax
          \global\tolerance=1000
          \global\spaceskip=0pt
          \global\xspaceskip=0pt
        \else
          \global\parfillskip=0pt plus 0.25\hsize\relax
          \global\emergencystretch=1.5em\relax
          \global\tolerance=400\relax
        \fi
      \else
        \global\parfillskip=0pt\relax
      \fi
    \else
      \global\parfillskip=0pt\relax
    \fi
  \else
    \global\parfillskip=0pt plus 1fil\relax
  \fi
}
\makeatother

\def\confName{CVPR}
\def\confYear{2027}
\makeatletter
\renewcommand\paragraph{\@startsection{paragraph}{4}{\z@}{0.5em \@plus 0.1em \@minus 0.1em}{-1em}{\normalfont\normalsize\bfseries}}
\makeatother
\title{Glance, then Select: Evidence-Tempered Visual Token Selection\\for Extremely Compressed Multimodal LLMs}
\author{%
Yuxuan Wang\textsuperscript{1},
Guangxiang Zhao\textsuperscript{2},
Yuxuan Tian\textsuperscript{1},
Yahui Han\textsuperscript{1},
Xinyu Tan\textsuperscript{1},\\
Feiyu Wang\textsuperscript{1},
Zongwei Lv\textsuperscript{1},
Xiangzheng Zhang\textsuperscript{2},
Lin Sun\textsuperscript{2},
Tong Yang\textsuperscript{1}\\[0.35em]
\textsuperscript{1}Peking University\qquad\textsuperscript{2}Qiyuan Tech%
}

\providecommand{\qwenlogo}{\raisebox{-0.32em}{\includegraphics[height=1.15em]{qwen-logo.pdf}}\hspace{0.32em}}
\providecommand{\gemmalogo}{\raisebox{-0.32em}{\includegraphics[height=1.15em]{gemma-logo.pdf}}\hspace{0.32em}}
\usepackage{dblfloatfix}
\makeatletter
\newif\ifinsupplement
\let\ltopd@addcontentsline\addcontentsline
\renewcommand{\addcontentsline}[3]{%
  \ltopd@addcontentsline{#1}{#2}{#3}%
  \ifinsupplement
    \def\ltopd@file{#1}%
    \def\ltopd@toc{toc}%
    \ifx\ltopd@file\ltopd@toc
      \ltopd@addcontentsline{stp}{#2}{#3}%
    \fi
  \fi
}
\newcommand{\printsupplementcontents}{%
  \section*{Contents of the Supplement}%
  \setcounter{tocdepth}{2}%
  \@starttoc{stp}%
}
\makeatother
\hypersetup{%
  pdftitle={Glance, then Select: Evidence-Tempered Visual Token Selection for Extremely Compressed Multimodal LLMs},%
  pdfauthor={Yuxuan Wang, Guangxiang Zhao, Yuxuan Tian, Yahui Han, Xinyu Tan, Feiyu Wang, Zongwei Lv, Xiangzheng Zhang, Lin Sun, Tong Yang}%
}

\begin{document}
\maketitle

\begin{abstract}
Multimodal large language models (MLLMs) encode a high-resolution image into thousands of visual tokens, which
dominate the cost of the prefill stage. Visual token reduction removes most of these tokens, but two problems remain at
extreme budgets. The vision encoder still processes every patch of the image, and the few surviving tokens often
miss the small regions that hold the answer. In this paper, we propose Glance-ETD, a two-stage token reduction
framework guided by a lightweight question-conditioned scorer. In the first stage, a glance encoder views the image
at low resolution and re-encodes only the question-relevant regions at native resolution. In the second stage,
Evidence-Tempered Diversity (ETD) selects the final tokens by combining diversity with the evidence predicted by the
scorer, so that the kept tokens cover the scene and retain the answer. A single-rollout self-distillation recipe trains the MLLM with the glance encoder. With only 5\% of the visual tokens,
Glance-ETD retains 91.6\% of the full-token performance of Qwen3.5-4B across fourteen benchmarks and outperforms the strongest prior method, LT-OPD, by 17.3 points and the same training recipe with DivPrune by 6.0. It also speeds up the time to first token on 4K images by 3.1$\times$ and
generalizes to three other MLLMs and to high-resolution document benchmarks,
with the largest gains on text-rich tasks.
\end{abstract}

\section{Introduction}
\label{sec:intro}

\begin{figure}[t]
\centering
\includegraphics[width=\linewidth]{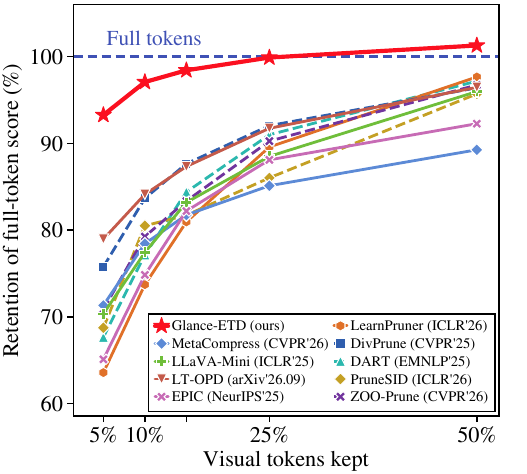}
\caption{Retention across visual-token budgets on Qwen3.5-4B: the mean over nine benchmarks\protect\footnotemark{} of each score divided by our non-thinking full-token score, with the same number of tokens per image. Solid lines are trained models\protect\footnotemark[2]{}; dashed lines are training-free and run on the released model, as in their papers. At 5\% of the tokens, Glance-ETD keeps 93.3\% (the mean of three training seeds; the other budgets use one seed), 17.5 points above DivPrune.}
\label{fig:budget}
\end{figure}
\footnotetext{V$^*$~\cite{wu2024vstar}, HR-Bench 4K~\cite{wang2025hrbench}, GQA~\cite{hudson2019gqa}, MMMU~\cite{yue2024mmmu}, MMBench (circular)~\cite{liu2024mmbench}, MME~\cite{fu2023mme}, POPE~\cite{li2023pope}, TextVQA~\cite{singh2019textvqa} and OCRBench~\cite{liu2024ocrbench}.}

Multimodal LLMs~\cite{liu2023llava,liu2024llava15,bai2025qwen25vl,qwen35} spend most of their prefill on visual tokens. A high-resolution image becomes thousands of
tokens, while the question about it is often answered in a few words. Visual token reduction~\cite{chen2024fastv,yang2025visionzip,alvar2025divprune} removes most of these
tokens before the language model, and LT-OPD~\cite{li2026ltopd} shows that a model trained on its own rollouts
against a full-token copy of itself can work with only a small fraction of them. Keeping one token in twenty is therefore a realistic operating point. We ask what limits a model at such an extreme budget and how to lift that limit.

\begin{figure*}[t]
\centering
\includegraphics[width=\textwidth]{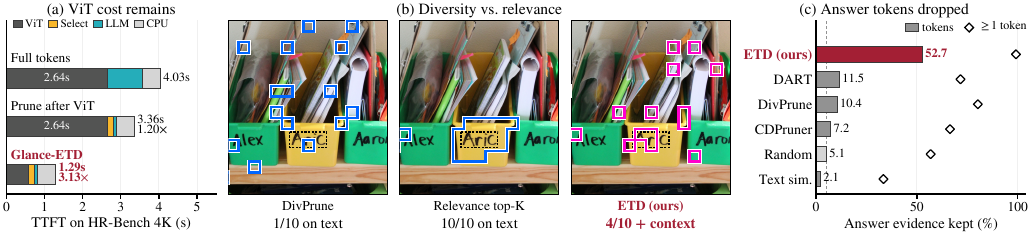}
\caption{Two problems at a strict 5\% visual-token budget on Qwen3.5-4B. (a) Time to first token on HR-Bench 4K for the full-token model, pruning after the ViT, and Glance-ETD (\cref{tab:latency}). Each bar is split into vision encoding, selection, LLM prefill and CPU preprocessing. Pruning after the ViT still spends 2.64\,s in the encoder; the glance encoder cuts that stage to 0.57\,s. (b) One TextVQA image at the same budget: outlines are kept tokens and the dotted box is the answer. DivPrune keeps 1 of 10 text tokens, relevance top-$K$ keeps all 10 but little context, and ETD keeps 4 plus context. (c) On 4{,}123 TextVQA validation questions with a model self-distilled at this budget, bars are the share of tokens overlapping the answer OCR box and diamonds the share of questions keeping at least one. The dashed line is the 5\% budget.}
\label{fig:motivation}
\end{figure*}

We find two limits (\cref{fig:motivation}). The first is cost. Pruning after the vision encoder shortens the
language model's prefill, but the encoder still attends over every patch. At high resolution the encoder becomes the
dominant stage, so discarding almost all tokens barely changes the time to the first token
(\cref{fig:motivation}a). Reducing tokens only pays off if the encoder also looks at less of the image.
\footnotetext[2]{The LT-OPD curve and \cref{tab:main}'s w/o merge rows turn off its learnable token merge; the released checkpoint keeps it, retains 80.5\% and is 11.1 points below Glance-ETD (\cref{tab:ltopd-variants}).}

The second limit is selection. When only one token in twenty survives, which tokens survive decides whether the
answer is still in the input. Diversity-based selectors such as DivPrune~\cite{alvar2025divprune} spread the budget over the scene. This works for questions about objects and layout but often drops the few tokens of a street sign or a phone number (\cref{fig:motivation}b,c). Selecting by relevance has the opposite failure: it finds the text but
loses the context that counting and existence questions need. Training does not remove this limit: a model self-distilled at this budget still loses most on reading benchmarks, and changing only the selector recovers much of that loss.

Both limits come down to one question: where is the evidence for this query? Knowing that, we could spend the encoder only on those regions and bias the selection toward them while still covering the scene. We build \textbf{Glance-ETD} around this idea. A small query-conditioned scorer reads the visual tokens
and the question and predicts an evidence map, and the same map guides two stages.

The first stage is a \emph{glance encoder}. It first glances at a low-resolution view and uses the scorer to decide which regions matter. Only those regions are then encoded at native resolution. The rest of the image keeps its coarse encoding, so the output stays on the native token grid and any
token selector can follow. The second stage is \emph{Evidence-Tempered Diversity} (ETD). It keeps the greedy diversity loop of DivPrune and tempers each candidate by its evidence. When the evidence map is flat, ETD behaves like
DivPrune; when the question points to a small region, the selection moves toward it while still covering the scene.
The scorer is trained once on box annotations with the MLLM frozen. A single-rollout self-distillation recipe then
trains the MLLM and the glance encoder together in 4.7 hours on eight A800 GPUs.

The two stages address the two limits directly. On Qwen3.5-4B with 5\% of the visual tokens, Glance-ETD retains
91.6\% of the full-token score over fourteen benchmarks, 17.3 points above the strongest prior method LT-OPD,
20.3 above DivPrune and 6.0 above our training recipe with DivPrune, and reaches the first token on 4K images 3.1$\times$ sooner than the full-token model.
It is the best method at every token budget (\cref{fig:budget}), with the largest margin at 5\%. The gains concentrate on reading and document questions where diversity alone fails. They also carry over to Gemma~4, GLM-4.6V and InternVL3.5 and to further high-resolution and document benchmarks.

In summary, our contributions are as follows.
\begin{itemize}
  \item We identify two limits of visual token reduction at extreme budgets. Pruning after the vision encoder leaves
    the encoder cost untouched, and training at the budget does not fix a selector that drops the answer.
  \item We propose Glance-ETD. Its glance encoder spends native resolution only on the regions that a
    question-conditioned scorer finds relevant. Its token selector ETD keeps the diversity loop of DivPrune and
    tempers it by the evidence from the same scorer.
  \item We run a controlled study at a strict 5\% budget over three training regimes, four MLLMs and 14
    benchmarks. The study measures the training recipe, the glance encoder and ETD one at a time and shows that the
    gains come mostly from reading and document questions.
\end{itemize}

\section{Related Work}
\label{sec:related}

\paragraph{Training-free token reduction.}
Most work on visual token reduction leaves the model weights unchanged. One line prunes tokens inside the language
model by their text-to-image attention, as in FastV~\cite{chen2024fastv}, SparseVLM~\cite{zhang2025sparsevlm} and
PyramidDrop~\cite{xing2025pdrop}. Another line selects or merges tokens before the language model using signals from
the encoder, as in VisionZip~\cite{yang2025visionzip} and PruMerge~\cite{shang2024prumerge}. DivPrune~\cite{alvar2025divprune} picks tokens
by max--min diversity and ignores the question.
CDPruner~\cite{zhang2025cdpruner} adds the question to the diversity objective. DART~\cite{wen2025dart} removes
duplicated tokens, and VisPruner~\cite{zhang2025vispruner} combines encoder attention with redundancy removal.
More recently, ZOO-Prune~\cite{kim2026zooprune} scores tokens by zeroth-order sensitivity, and
PruneSID~\cite{fang2026prunesid} removes redundancy within groups of similar tokens. VisionTrim~\cite{yu2026visiontrim}
complements attention-based selection with text-guided tokens, and HAWK~\cite{zhu2026hawk} weights text-to-image
attention by head. TRIM~\cite{song2025trim} scores tokens by CLIP similarity to the question, and
ATP-LLaVA~\cite{ye2025atpllava} and Dynamic-LLaVA~\cite{huang2025dynamicllava} learn pruning ratios per layer. SPARE~\cite{lee2026spare} measures diversity by subspace reconstruction, and
MetaCompress~\cite{wang2026metacompress} learns a prompt-agnostic compression matrix. All of
them act after the vision encoder and leave its cost unchanged. Inside the ViT, token pruning and
merging~\cite{rao2021dynamicvit,liang2022evit,bolya2023tome} and coarse-to-fine or mixed-resolution
encoding~\cite{wang2020gfnet,chen2023cfvit,ronen2023quadformer} were designed for classification. Mini-Gemini~\cite{li2024minigemini} and FocusLLaVA~\cite{zhu2024focusllava} bring
coarse-to-fine to MLLMs after a dense encoder pass; the glance encoder makes that pass itself cheaper.
FastVLM~\cite{vasu2025fastvlm} replaces the encoder with a hybrid one that emits fewer tokens at high resolution.

\paragraph{Training with fewer visual tokens.}
Training can recover much of the accuracy lost at small budgets. Q-Former~\cite{li2023blip2} and Matryoshka
Multimodal Models~\cite{cai2025m3} learn compact or nested sets of visual tokens. LLaVA-Mini~\cite{zhang2025llavamini} compresses the
image into a few query tokens before the language model. EPIC~\cite{wen2025epic} distills a compressed student from a
full-token teacher step by step. LearnPruner~\cite{takezoe2026learnpruner} learns a pruning module after the vision
encoder and pairs it with text-guided pruning inside the language model. On-policy distillation~\cite{hinton2015kd,agarwal2024gkd,gu2024minillm} trains a student on its own samples
under token-level teacher supervision. LT-OPD~\cite{li2026ltopd} is closest to
our setting: it distills a student on its own rollouts from a frozen full-token copy, with a curriculum over the
budget. We adopt its recipe and study the selector and the vision encoder on top of it.

\paragraph{Evidence in high-resolution images.}
On text-rich images the answer often sits in a few small regions. ET-Prune~\cite{ding2026etprune} reads
question-conditioned evidence from decoder attention, protects text-like regions and sets a budget per sample. It
prunes inside the LLM and keeps about half of the tokens. DocPrune~\cite{choi2026docprune} removes background and
irrelevant blocks from document images. HiRED~\cite{arif2025hired} splits the token budget across
high-resolution partitions by encoder attention. ETD works at 5\% of the tokens before the LLM. Its evidence comes from a
scorer trained on Visual CoT~\cite{shao2024viscot} boxes, and its diversity term keeps the scene covered. V*~\cite{wu2024vstar} and divide-and-conquer
systems~\cite{wang2025hrbench} search a high-resolution image over several steps. ViCrop~\cite{zhang2025vicrop} crops and FlexAttention~\cite{li2024flexattention} selects high-resolution tokens where the MLLM attends,
ZoomEye~\cite{shen2025zoomeye} searches a zoom tree, DeepEyes~\cite{zheng2026deepeyes} learns to zoom with
reinforcement learning, S$^2$~\cite{shi2024s2} encodes several scales of the image, and PS3~\cite{shi2025ps3}, the encoder behind VILA-HD, is pre-trained to encode
a low-resolution view and then only the high-resolution patches the prompt selects. The glance encoder
reuses the MLLM's own encoder, adapted by self-distillation, and makes its single pass cheaper on large images.

\section{Method}
\label{sec:method}

\begin{figure*}[t]
\centering
\includegraphics[width=\linewidth]{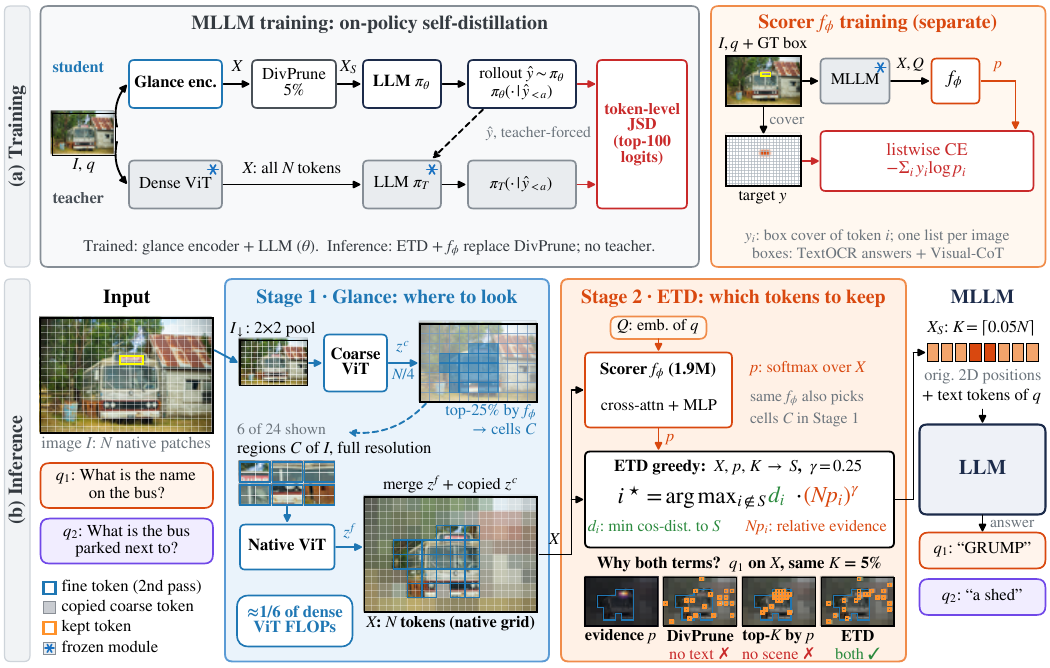}
\caption{\textbf{Glance-ETD.} (a) Training: the student (glance encoder and LLM) selects with DivPrune, rolls out,
and is distilled token by token towards a frozen full-token teacher; the scorer $f_\phi$ is trained separately on
box targets with the MLLM frozen. (b) Inference: Stage~1 (\cref{sec:glance}) encodes the image at a quarter of the
patches, the scorer $f_\phi$ picks the 25\% of the coarse cells with the highest question relevance, and only these
are re-encoded at native resolution, giving a grid $X$ with all $N$ native positions. Stage~2 (\cref{sec:etd}) maps
$X$ and the question to an evidence map $p$, and ETD greedily selects $K$ tokens of $X$ by $d_i(Np_i)^{0.25}$ (shown for
$q_1$, on an illustrative evidence map).}
\label{fig:pipeline}
\end{figure*}

Glance-ETD reduces the visual input in two stages (\cref{fig:pipeline,alg:glance-etd}): the glance encoder
decides where to look at full resolution (\cref{sec:glance}), and ETD chooses the $K$ tokens passed to the language
model (\cref{sec:etd}). One small question-conditioned scorer drives both decisions. We then describe the teacher-KL
diagnostic (\cref{sec:kl}) and how the two stages are trained (\cref{sec:training}).

\subsection{Preliminaries}
\label{sec:prelim}

\paragraph{Problem setup.} An image is encoded into $N$ visual tokens $X=\{x_i\}_{i=1}^{N}$ in the LLM embedding
space, after the patch merger. Given a question $q$ and a budget $K\ll N$, a selector returns an index set
$S\subset\{1,\dots,N\}$ with $|S|=K$, and only $X_S$ is fed to the language model, with its original positions. We
use two budget rules. \emph{Strict}: $K=\lceil rN\rceil$ with $r=5\%$, so that $\sum K/\sum N=5.0\%$ on every
benchmark. \emph{Floor-32}: $K=\max(32,\lceil 0.05N\rceil)$, the rule of the LT-OPD release, which keeps 7.1\% of
all tokens because small images hit the floor. All selectors compared in this paper keep the same $K$ per image and
differ only in which tokens they keep, not in how many visual tokens each image receives.

\paragraph{Diversity selection.} DivPrune~\cite{alvar2025divprune} starts from the token least similar to the mean
token and greedily adds
\begin{equation}
  i^\star=\arg\max_{i\notin S}\; d_i,\qquad d_i=\min_{j\in S}\big(1-\cos(x_i,x_j)\big),
  \label{eq:divprune}
\end{equation}
until $|S|=K$. The max--min rule, the greedy $k$-center heuristic~\cite{gonzalez1985kcenter}, spreads the budget over the image and ignores $q$. ETD keeps it as its diversity
term.

\paragraph{On-policy self-distillation.} LT-OPD~\cite{li2026ltopd} trains a student MLLM that sees only $X_S$ to
match a frozen copy of itself that sees all of $X$. The student generates answers with the pruned input, and the
teacher provides token-level supervision along these rollouts, as in on-policy
distillation~\cite{agarwal2024gkd,gu2024minillm}. We adopt this objective for every model trained in this paper as well.

\subsection{Stage 1: Glance Encoder}
\label{sec:glance}

Post-encoder selection leaves the ViT encoder~\cite{dosovitskiy2021vit} untouched, and for large images the encoder dominates: on HR-Bench 4K,
89\% of its FLOPs are global attention over about 50k patches. The glance encoder decides, before full-resolution
encoding, which parts of the image the encoder should see in full detail.

\paragraph{Two-pass encoding.} (i) \emph{Glance.} The image is average-pooled $2\times2$ and encoded with full
attention at a quarter of the patches, giving one coarse token per $2\times2$ cell of the native merged grid.
(ii) \emph{Look.} The evidence scorer of \cref{sec:scorer} reads the coarse tokens and the question, and the 25\% of
the cells with the highest relevance logits $r$ are selected; the native patches of all selected cells are encoded as one sequence with global attention among them, with the absolute and rotary
position codes of their native positions. (iii) \emph{Merge.} After the patch merger, selected cells take their
second-pass tokens and every other cell copies its coarse token to its four native slots. The grid has the size of
the dense encoder's, so any post-encoder selector applies unchanged, and the LLM gives every slot its own M-RoPE~\cite{wang2024qwen2vl}
position. The four copies are identical features, so once one is kept the others have $d_i=0$ and are not selected
again. The encoder costs about a sixth of the dense FLOPs,
and since the 5\% budget is a fifth of the high-resolution pool, most kept tokens come from native-resolution regions.

\paragraph{Training.} The glance encoder is trained inside the MLLM by the self-distillation objective alone. During training the cells are
chosen by max--min diversity (\cref{eq:divprune}), so the MLLM never depends on the scorer; the cell choice is
discrete and receives no gradient. At inference the scorer chooses the cells. It is applied to images
with at least 4096 patches; smaller images are encoded densely at their native resolution.

\subsection{Stage 2: Evidence-Tempered Diversity}
\label{sec:etd}

ETD keeps the diversity loop of \cref{eq:divprune} and tilts it towards query evidence. It has two parts: a scorer
that produces an evidence map, and a greedy rule that uses it.

\paragraph{Evidence scorer.}
\label{sec:scorer}
A small network $f_\phi$ reads the visual tokens and the question embeddings and outputs a relevance logit $r_i$ for
every visual token. Visual tokens are projected to $h=256$ dimensions and given a 2D
sine--cosine position code; one cross-attention layer lets them attend to the projected question tokens, producing a
context $c_i$; an MLP on $[h_i,\,c_i,\,h_i\odot\bar q]$, with $\bar q$ the mean question feature, produces
$r_i$. The scorer has about 1.9M parameters and reads the MLLM's own token features, so no extra encoder is
needed; with the glance encoder it reads the coarse grid in Stage~1 and the merged native grid in Stage~2, the same
input that ETD selects from, at about 2\,ms per call. The evidence distribution is $p=\mathrm{softmax}(r)$.

\paragraph{Selection rule.} ETD replaces the greedy score of \cref{eq:divprune}\linebreak
with the evidence-tempered score below,
\begin{equation}
  i^\star=\arg\max_{i\notin S}\; d_i\,\big(N p_i\big)^{\gamma},\qquad \gamma=0.25.
  \label{eq:etd}
\end{equation}
The factor $Np_i$ is the evidence of token $i$ relative to a uniform map, and $\gamma$ is an inverse temperature. If
the scorer finds no evidence ($p$ uniform), every weight is one and ETD is exactly DivPrune. A peaked map on a reading
question moves the selection to the evidence; a broad map on a scene question leaves it close to diversity. No
threshold or question classifier is involved. The value $\gamma=0.25$ gives the lowest teacher KL on training
questions (\cref{sec:kl}) and is used unchanged for every model and budget. Because the distance term remains, a text region does not
absorb the whole budget: once a few tokens of a word are kept, its neighbours have small $d_i$ and the budget moves on.
Hard splits, gates and a per-question $\gamma$ are compared in \cref{sec:ablation}. We compute \cref{eq:etd} in log
space with the weights shifted by their maximum, which keeps the ranking exact.

\paragraph{Scorer supervision.} The scorer is trained from boxes, not from benchmark labels. For TextVQA training
questions we align answer words to TextOCR word boxes; for Visual-CoT~\cite{shao2024viscot} we use the annotated
evidence boxes, which cover both text and general objects. The target $y_i$ of token $i$ is the fraction of its cell
covered by evidence boxes, normalised to $\sum_i y_i=1$ over the image. The loss is listwise: each image is one list,
and the scorer minimises the cross-entropy $-\sum_i y_i\log p_i$ over all its $N$ tokens, so only the ranking of $r$
within an image is supervised.
Some images overlap evaluation sets (\eg 105 of 1{,}284 DocVQA images); a scorer trained without them is
in \cref{sec:supp-dedup}. The scorer reads the frozen MLLM's tokens of the densely encoded image and its question embeddings, and is trained
with AdamW~\cite{loshchilov2019adamw} ($3\times10^{-4}$, 16 images per step) for three epochs of 60K images sampled from 422K records (28K TextVQA,
the rest Visual-CoT, sampled at 0.3 of the TextVQA weight).

\subsection{Teacher-KL Diagnostic}
\label{sec:kl}

To compare selection rules without running benchmarks, we measure how far a subset moves the model's predictions away from the full-token ones. For a full-token greedy answer $A$ of at most 32 tokens, the cost of a subset $S$ is
\begin{equation}
  \mathcal{L}(S)=\sum_{a=1}^{|A|}\KL\!\big(\pi(\cdot\mid X,q,A_{<a})\,\big\|\,\pi(\cdot\mid X_S,q,A_{<a})\big),
  \label{eq:kl}
\end{equation}
computed with teacher forcing along $A$ on the untrained model. On 8{,}000 training questions at a fixed $K$, ETD reaches 0.46 of the $\mathcal{L}$ of DivPrune, almost all of the reduction on text questions; among exponents $\gamma\in\{0.25,0.5,1,2,4,8\}$, 0.25 gives the lowest $\mathcal{L}$. Fitting $\mathcal{L}$ as a power law in the kept fraction gives a per-image budget $K_m\propto N_m^{0.45}$ (\cref{sec:supp-kl,sec:alloc}; not used in our main results).

\subsection{Training and Inference}
\label{sec:training}

\paragraph{Training recipe.} The training recipe is the third part of Glance-ETD. We train Qwen3.5-4B by on-policy
self-distillation (\cref{sec:prelim}): the student rolls out with 5\% of the tokens, chosen by DivPrune, and merges
the discarded tokens into the kept ones. The loss is the token-level generalised Jensen--Shannon
divergence ($\alpha=0.5$) between teacher and student along the student's rollouts, on a top-100 token
support with an aggregated tail bucket. One rollout per prompt and a 256-token answer cap train in 4.7\,h on
8 A800 GPUs instead of 17.4\,h at the same accuracy (\cref{sec:supp-eff}). We run 175 steps on a 14K-prompt
self-distillation set~\cite{li2026ltopd}, of which 1.5K prompts are replaced by spatial-reasoning samples. The vision
encoder needs a learning rate of $2\times10^{-6}$, ten times the usual rate for self-distillation, to adapt to its
two-resolution input: the higher rate raises the dense model with DivPrune from 85.3\% to 88.8\% and Glance-ETD by
3.4 points. \Cref{sec:components} reports the recipe and the two stages one at a time, at both rates.

\paragraph{Scorer and inference.} The scorer is trained once on the features of the untrained MLLM and shared by all
Qwen models; retraining it on the features of each trained model does not help (\cref{sec:supp-dedup}). At
inference, the scorer chooses the glance cells, the glance encoder produces the token grid, the scorer computes $p$
on it, and ETD selects $K$ tokens, which enter the LLM with their original positions. The MLLM is thus trained with
diversity selection and evaluated with the scorer. This keeps the MLLM independent of the scorer, so the same rule
applies to the untrained model; training with ETD instead is compared in \cref{sec:supp-vit}.

\section{Experiments}
\label{sec:exp}

\begin{table*}[t]
\centering
\footnotesize
\setlength{\tabcolsep}{2.2pt}
\caption{Comparison with token-reduction methods on Qwen3.5-4B and Gemma~4 E4B-it at a strict 5\% budget, where every method keeps the same number $K$ of visual tokens per image. Methods are grouped into training-free methods, learned modules trained with the MLLM frozen (ETD trains only its scorer), and methods that adapt the MLLM. The two self-distillation rows share the same training recipe and differ only in the token selector; Self-distillation + ETD is our method with the dense encoder. Ret.\ is the score on each benchmark relative to the same model's full-token score, averaged over the fourteen benchmarks. Self-distilled rows report the mean $\pm$ std over three training seeds (\cref{tab:seeds}). $^\ddagger$GQA answers are lower-cased before scoring.}
\label{tab:main}
\resizebox{\textwidth}{!}{%
\begin{tabular}{lccccc|ccccc|cccc|c}
\toprule
 & \multicolumn{5}{c|}{General} & \multicolumn{5}{c|}{Fine-grained high resolution} & \multicolumn{4}{c|}{Text and document} & Ret.\,(\%) \\
\cmidrule(lr){2-6}\cmidrule(lr){7-11}\cmidrule(lr){12-15}
Method & GQA & MMMU & MMB & MME & POPE & V$^*$ & HR-4K & HR-8K & MME-RW & HRScene & TextVQA & OCRB & DocVQA & InfoVQA & All \\
\midrule
\qwenhead{Training-free} \\
Full tokens (100\%) & 60.0 & 51.3 & 79.0 & 2218 & 89.9 & 80.6 & 78.0 & 72.6 & 48.9 & 62.5 & 80.5 & 848 & 94.6 & 79.4 & 100.0 \\
DivPrune~\cite{alvar2025divprune} \venue{CVPR'25} & 45.3 & 46.3 & 63.4 & 1695 & 77.4 & 70.2 & 73.4 & 70.5 & 38.1 & 49.6 & 49.2 & 264 & 32.1 & 21.7 & 71.3 \\
DART~\cite{wen2025dart} \venue{EMNLP'25} & 45.0 & 48.2 & 53.6 & 1529 & 77.6 & 49.7 & 61.5 & 60.4 & 30.1 & 45.0 & 49.6 & 125 & 24.9 & 21.8 & 62.8 \\
CDPruner~\cite{zhang2025cdpruner} \venue{NeurIPS'25} & 35.7 & 47.2 & 59.7 & 1605 & 74.3 & 53.9 & 62.7 & 56.4 & 29.7 & 43.3 & 43.3 & 215 & 21.8 & 20.5 & 61.8 \\
PruneSID~\cite{fang2026prunesid} \venue{ICLR'26} & 44.3 & 47.0 & 55.0 & 1602 & 77.9 & 69.6 & 73.8 & 68.8 & 35.0 & 49.2 & 25.1 & 108 & 22.3 & 20.7 & 65.2 \\
VisionTrim~\cite{yu2026visiontrim} \venue{ICLR'26} & 41.8 & 45.8 & 55.4 & 1514 & 69.4 & 48.7 & 63.8 & 59.0 & 23.9 & 43.8 & 20.3 & 89 & 15.8 & 21.5 & 56.9 \\
HAWK~\cite{zhu2026hawk} \venue{CVPR'26} & 30.7 & 47.4 & 49.0 & 1336 & 70.9 & 49.2 & 60.4 & 56.2 & 22.5 & 44.3 & 16.5 & 72 & 9.7 & 18.9 & 52.9 \\
ZOO-Prune~\cite{kim2026zooprune} \venue{CVPR'26} & 46.2 & 46.4 & 60.1 & 1646 & 81.3 & 72.3 & 72.2 & 67.9 & 36.8 & 49.4 & 26.8 & 118 & 25.5 & 20.8 & 67.0 \\
\midrule
\qwenhead{Frozen MLLM, learned module} \\
LLaVA-Mini~\cite{zhang2025llavamini} \venue{ICLR'25} & 43.5 & 47.9 & 65.3 & 1579 & 75.7 & 56.5 & 68.6 & 61.7 & 32.4 & 50.1 & 34.9 & 236 & 34.7 & 27.4 & 66.8 \\
LearnPruner~\cite{takezoe2026learnpruner} \venue{ICLR'26} & 43.3 & 45.6 & 55.1 & 1505 & 72.1 & 48.2 & 69.0 & 60.9 & 29.2 & 43.7 & 28.2 & 87 & 12.7 & 19.3 & 58.8 \\
MetaCompress~\cite{wang2026metacompress} \venue{CVPR'26} & 41.9 & 46.9 & 66.1 & 1576 & 69.8 & 63.4 & 70.4 & 62.7 & 27.8 & 49.2 & 41.2 & 241 & 28.5 & 22.9 & 65.9 \\
\rowcolor{gray!12}ETD (ours) & 47.2 & 46.1 & 63.2 & 1772 & 84.8 & 78.0 & 75.8 & 73.8 & 42.3 & 52.6 & 70.2 & 426 & 71.9 & 36.8 & \textbf{82.0} \\
\midrule
\qwenhead{MLLM adaptation} \\
EPIC$^\ddagger$~\cite{wen2025epic} \venue{NeurIPS'25} & 42.1 & 44.4 & 57.0 & 1233 & 66.3 & 54.5 & 64.8 & 59.8 & 39.8 & 48.2 & 50.9 & 120 & 25.4 & 22.9 & 63.0 \\
LT-OPD~\cite{li2026ltopd} \venue{arXiv'26.09} w/o merge & 47.4 & 48.6 & 66.4 & 1771 & 81.9 & 67.0 & 69.6 & 65.1 & 40.5 & 51.4 & 55.8 & 341 & 40.6 & 25.6 & 74.3 \\
Self-distillation + DivPrune & 54.2 & 48.3 & 73.5 & 1988 & 85.9 & 75.0 & 77.1 & 73.1 & 45.5 & 55.3 & 70.4 & 489 & 67.2 & 35.7 & 85.6$\pm$0.2 \\
\rowcolor{gray!12}\textbf{Self-distillation + ETD (ours)} & 55.0 & 48.8 & 74.9 & 2009 & 88.1 & 79.1 & 77.2 & 73.2 & 45.6 & 56.1 & 76.3 & 597 & 83.5 & 49.6 & 90.5$\pm$0.4 \\
\rowcolor{gray!12}\textbf{Glance-ETD (ours)} & 54.7 & 49.6 & 75.2 & 1985 & 88.2 & 83.6 & 78.5 & 75.9 & 48.8 & 56.7 & 76.3 & 590 & 82.6 & 48.1 & \textbf{91.6}$\pm$0.3 \\
\midrule
\gemmahead{Training-free} \\
Full tokens (100\%) & 41.3 & 46.9 & 73.6 & 1686 & 88.1 & 58.1 & 61.1 & 54.4 & 42.4 & 54.6 & 71.9 & 764 & 88.3 & 62.0 & 100.0 \\
DivPrune~\cite{alvar2025divprune} \venue{CVPR'25} & 37.1 & 42.9 & 60.5 & 1103 & 83.9 & 49.2 & 53.0 & 46.1 & 23.7 & 43.7 & 38.9 & 184 & 14.7 & 15.0 & 66.8 \\
DART~\cite{wen2025dart} \venue{EMNLP'25} & 33.2 & 41.3 & 52.5 & 934 & 72.4 & 34.6 & 40.2 & 35.8 & 17.9 & 34.7 & 19.9 & 83 & 9.1 & 12.3 & 53.1 \\
CDPruner~\cite{zhang2025cdpruner} \venue{NeurIPS'25} & 35.3 & 42.2 & 54.7 & 962 & 78.4 & 28.8 & 44.4 & 38.6 & 14.3 & 37.3 & 28.9 & 138 & 10.4 & 13.5 & 55.9 \\
PruneSID~\cite{fang2026prunesid} \venue{ICLR'26} & 36.8 & 41.1 & 60.2 & 1164 & 79.9 & 27.2 & 41.2 & 38.4 & 21.4 & 39.7 & 30.9 & 246 & 12.3 & 16.1 & 60.1 \\
VisionTrim~\cite{yu2026visiontrim} \venue{ICLR'26} & 33.5 & 38.4 & 55.7 & 1046 & 78.0 & 41.4 & 54.4 & 47.5 & 22.7 & 39.6 & 24.0 & 138 & 9.6 & 15.6 & 60.7 \\
HAWK~\cite{zhu2026hawk} \venue{CVPR'26} & 30.4 & 40.2 & 50.4 & 874 & 69.2 & 21.5 & 30.3 & 26.9 & 10.8 & 31.3 & 9.6 & 85 & 7.4 & 13.6 & 45.1 \\
ZOO-Prune~\cite{kim2026zooprune} \venue{CVPR'26} & 36.5 & 43.1 & 60.1 & 1062 & 81.8 & 34.0 & 41.2 & 36.6 & 19.9 & 39.9 & 24.9 & 151 & 10.6 & 15.2 & 58.7 \\
\midrule
\gemmahead{Frozen MLLM, learned module} \\
LLaVA-Mini~\cite{zhang2025llavamini} \venue{ICLR'25} & 30.3 & 39.9 & 30.5 & 1018 & 65.5 & 38.7 & 32.5 & 35.4 & 29.2 & 34.9 & 11.9 & 62 & 8.5 & 18.3 & 51.1 \\
LearnPruner~\cite{takezoe2026learnpruner} \venue{ICLR'26} & 32.6 & 40.9 & 47.9 & 952 & 72.7 & 28.3 & 36.3 & 37.3 & 14.6 & 35.3 & 13.3 & 25 & 4.1 & 12.0 & 49.4 \\
MetaCompress~\cite{wang2026metacompress} \venue{CVPR'26} & 35.6 & 40.6 & 59.4 & 1063 & 77.7 & 27.2 & 33.9 & 31.2 & 17.6 & 36.8 & 22.9 & 196 & 10.8 & 16.5 & 55.0 \\
\rowcolor{gray!12}ETD (ours) & 37.7 & 42.2 & 63.3 & 1186 & 83.3 & 55.0 & 62.0 & 56.0 & 31.8 & 44.0 & 59.9 & 396 & 39.1 & 21.0 & \textbf{78.6} \\
\midrule
\gemmahead{MLLM adaptation} \\
EPIC~\cite{wen2025epic} \venue{NeurIPS'25} & 40.5 & 43.1 & 55.3 & 1362 & 78.2 & 44.5 & 47.6 & 42.2 & 34.5 & 41.5 & 24.5 & 171 & 13.7 & 17.9 & 66.1 \\
LT-OPD~\cite{li2026ltopd} \venue{arXiv'26.09} w/o merge & 39.3 & 43.1 & 63.5 & 1176 & 82.5 & 37.7 & 52.2 & 47.8 & 27.5 & 42.6 & 39.1 & 349 & 16.2 & 16.1 & 68.7 \\
Self-distillation + DivPrune & 41.3 & 42.9 & 70.8 & 1386 & 87.3 & 49.7 & 57.3 & 53.0 & 29.3 & 44.7 & 56.8 & 440 & 26.4 & 18.6 & 78.1$\pm$0.8 \\
\rowcolor{gray!12}\textbf{Self-distillation + ETD (ours)} & 41.2 & 42.7 & 71.2 & 1459 & 86.2 & 50.4 & 60.9 & 55.5 & 35.3 & 46.8 & 66.1 & 546 & 42.7 & 23.6 & \textbf{84.2}$\pm$0.7 \\
\rowcolor{gray!12}\textbf{Glance-ETD (ours)} & 40.8 & 41.4 & 70.0 & 1472 & 86.1 & 52.9 & 61.6 & 51.0 & 35.2 & 47.9 & 64.7 & 542 & 40.0 & 25.0 & 83.6$\pm$1.4 \\
\bottomrule
\end{tabular}
}
\end{table*}

\subsection{Experimental Setup}
\label{sec:setup}

\paragraph{Models.} The main experiments use Qwen3.5-4B~\cite{qwen35} in non-thinking mode, and \cref{sec:main}
repeats the comparison on three other MLLMs. \textbf{Glance-ETD} is the self-distilled model with the glance
encoder of \cref{sec:training}, evaluated with scorer-chosen glance cells and ETD. The \emph{dense-encoder
variant} is trained the same way without the glance encoder and evaluated with ETD, and \emph{untrained} is
the released model.

\paragraph{Benchmarks and metric.} \Cref{tab:main} covers fourteen benchmarks: nine standard ones,
V$^*$~\cite{wu2024vstar}, HR-Bench 4K~\cite{wang2025hrbench}, GQA~\cite{hudson2019gqa},
MMMU~\cite{yue2024mmmu}, circular MMBench~\cite{liu2024mmbench}, MME~\cite{fu2023mme},
POPE~\cite{li2023pope}, TextVQA~\cite{singh2019textvqa} and OCRBench~\cite{liu2024ocrbench}, and five with
larger images, HR-Bench 8K~\cite{wang2025hrbench},
MME-RealWorld~\cite{zhang2025mmerealworld}, HRScene~\cite{zhang2025hrscene}, DocVQA~\cite{mathew2021docvqa}
and InfoVQA~\cite{mathew2022infovqa}. We summarise a model by its
\emph{retention} over all fourteen, written Ret.\ All: the mean ratio of each score to the same model on full
tokens, the first row of each block and repeated in \cref{sec:supp-proto}. Ablations and budget sweeps use the
retention over the nine.

\paragraph{Budgets and baselines.} The main budget is the strict 5\% of \cref{sec:prelim}, and floor-32 is in
\cref{sec:floor32}. The baselines are peer-reviewed methods from 2025--2026, re-run on Qwen3.5-4B with the same
$K$ per image as ETD and ported to select after the vision merger; \cref{sec:supp-bl} gives the ports.
LLaVA-Mini, LearnPruner and MetaCompress train their modules on the same 14K prompts with the MLLM frozen.
LT-OPD w/o merge retrains that recipe on each model without the token merge, which our trained models keep
(\cref{sec:supp-ltopd}).

\addtocounter{table}{1}
\begin{table*}[t]
\centering
\small
\setlength{\tabcolsep}{3.2pt}
\caption{Ablation of the design choices in Glance-ETD on Qwen3.5-4B and Gemma~4 E4B-it at a strict 5\% budget. Each row changes one choice of the full pipeline: Stage~1 rows change which glance cells are re-encoded, and Stage~2 rows change how tokens are selected. Ret.\ is the retention over the nine LT-OPD benchmarks (six of which are shown), and $\Delta$ is its change from the full pipeline. All rows use training seed 0; three-seed means of seven Qwen rows keep the same order (\cref{sec:supp-abl}).}
\label{tab:ablation}
\resizebox{\textwidth}{!}{%
\begin{tabular}{lcccccccc|cccccccc}
\toprule
 & \multicolumn{8}{c|}{\qwenlogo Qwen3.5-4B, Glance-ETD} & \multicolumn{8}{c}{\gemmalogo Gemma~4 E4B-it, Glance-ETD} \\
\cmidrule(lr){2-9}\cmidrule(lr){10-17}
Setting & V$^*$ & HR-4K & GQA & POPE & TVQA & OCRB & Ret. & $\Delta$ & V$^*$ & HR-4K & GQA & POPE & TVQA & OCRB & Ret. & $\Delta$ \\
\midrule
\rowcolor{gray!12}\textbf{Full pipeline} & 82.7 & 78.9 & 53.8 & 88.2 & 76.8 & 583 & 92.9 & -- & 51.8 & 63.3 & 43.6 & 86.2 & 65.6 & 550 & 92.9 & -- \\
\midrule
\multicolumn{17}{l}{\textit{Stage 1: which cells to re-encode}} \\
max--min diversity & 77.5 & 76.2 & 53.9 & 87.9 & 76.4 & 580 & 91.9 & -1.0 & 47.1 & 58.0 & 43.7 & 86.0 & 62.1 & 554 & 90.8 & -2.1 \\
random cells & 73.8 & 74.1 & 53.9 & 88.1 & 76.8 & 585 & 90.9 & -2.0 & 47.6 & 55.9 & 43.6 & 86.5 & 61.8 & 580 & 90.9 & -2.0 \\
\midrule
\multicolumn{17}{l}{\textit{Stage 2: which tokens (same $K$ per image)}} \\
w/o evidence: DivPrune & 77.5 & 79.6 & 53.3 & 86.7 & 70.8 & 502 & 90.0 & -2.9 & 51.8 & 58.6 & 43.5 & 87.2 & 60.4 & 467 & 89.7 & -3.2 \\
strong evidence bias: $\gamma=64$ & 77.5 & 78.6 & 51.7 & 86.7 & 73.6 & 523 & 89.3 & -3.6 & 49.2 & 58.1 & 43.8 & 86.8 & 61.1 & 515 & 90.4 & -2.5 \\
$\gamma=1$: $d_i\,Np_i$ & 84.8 & 78.4 & 52.6 & 87.8 & 75.8 & 578 & 92.0 & -0.9 & 52.4 & 59.5 & 42.8 & 84.9 & 63.6 & 536 & 90.9 & -2.0 \\
$\gamma=0.125$ & 78.0 & 79.2 & 54.5 & 87.7 & 76.4 & 584 & 92.4 & -0.5 & 50.3 & 61.7 & 44.0 & 86.5 & 66.4 & 536 & 91.9 & -1.0 \\
$\gamma$ set per question by the text confidence $t$ & 80.6 & 80.2 & 54.2 & 87.8 & 76.2 & 579 & 92.2 & -0.7 & 51.3 & 59.8 & 43.0 & 85.9 & 63.8 & 540 & 91.5 & -1.4 \\
hard split: top-$K/2$ by $p$ + DivPrune & 81.7 & 78.6 & 53.4 & 86.5 & 75.0 & 541 & 91.3 & -1.6 & 50.8 & 60.4 & 43.5 & 86.8 & 62.6 & 514 & 90.3 & -2.6 \\
hard gate: split only if $t>0.5$ & 77.5 & 79.2 & 53.3 & 86.5 & 74.7 & 539 & 90.8 & -2.1 & 52.4 & 60.9 & 43.1 & 86.7 & 62.2 & 512 & 91.1 & -1.8 \\
scorer given a wrong question & 80.6 & 78.8 & 52.5 & 87.7 & 72.7 & 523 & 90.7 & -2.2 & 51.8 & 59.2 & 43.2 & 85.6 & 64.2 & 504 & 90.8 & -2.1 \\
DPP-$r$ instead of ETD & 82.7 & 78.9 & 49.3 & 87.7 & 76.5 & 592 & 91.6 & -1.3 & 49.2 & 60.9 & 43.2 & 85.6 & 66.2 & 554 & 91.5 & -1.4 \\
\bottomrule
\end{tabular}}
\end{table*}

\addtocounter{table}{-2}
\begin{table}[t]
\centering
\footnotesize
\setlength{\tabcolsep}{3pt}
\caption{Component ablation on Qwen3.5-4B at a strict 5\% budget. Glance: glance encoder, otherwise dense. ETD: evidence-tempered selection, otherwise DivPrune. The 10$\times$ and 1$\times$ LR groups self-distil the MLLM with that ViT learning rate (1$\times$ is the LT-OPD recipe); No train is the released model. Ret.: retention over the nine LT-OPD benchmarks, mean $\pm$ std over three seeds for trained models. TFLOPs: ViT + LLM prefill on HR-Bench 4K. Shaded: Glance-ETD.}
\label{tab:components}
\resizebox{\columnwidth}{!}{%
\begin{tabular}{@{}lcccccccccc@{}}
\toprule
& \multicolumn{4}{c}{10$\times$ LR} & \multicolumn{4}{c}{1$\times$ LR} & \multicolumn{2}{c}{No train} \\
\cmidrule(lr){2-5}\cmidrule(lr){6-9}\cmidrule(lr){10-11}
Glance & -- & -- & \checkmark & \cellcolor{gray!12}\checkmark & -- & -- & \checkmark & \checkmark & -- & -- \\
ETD & -- & \checkmark & -- & \cellcolor{gray!12}\checkmark & -- & \checkmark & -- & \checkmark & -- & \checkmark \\
\midrule
V$^*$ & 75.0 & 79.1 & 74.9 & \cellcolor{gray!12}83.6 & 73.8 & 76.4 & 73.1 & 82.0 & 70.2 & 78.0 \\
HR-4K & 77.1 & 77.2 & 76.8 & \cellcolor{gray!12}78.5 & 77.2 & 77.2 & 77.1 & 78.3 & 73.4 & 75.8 \\
GQA & 54.2 & 55.0 & 54.1 & \cellcolor{gray!12}54.7 & 52.3 & 53.3 & 52.0 & 52.5 & 45.3 & 47.2 \\
MMMU & 48.3 & 48.8 & 50.4 & \cellcolor{gray!12}49.6 & 48.9 & 48.9 & 48.3 & 48.3 & 46.3 & 46.1 \\
MMB & 73.5 & 74.9 & 73.8 & \cellcolor{gray!12}75.2 & 70.6 & 71.8 & 70.9 & 72.1 & 63.4 & 63.2 \\
MME & 1988 & 2009 & 2001 & \cellcolor{gray!12}1985 & 1865 & 1843 & 1856 & 1840 & 1695 & 1772 \\
POPE & 85.9 & 88.1 & 86.2 & \cellcolor{gray!12}88.2 & 85.2 & 87.8 & 85.2 & 87.4 & 77.4 & 84.8 \\
TextVQA & 70.4 & 76.3 & 70.5 & \cellcolor{gray!12}76.3 & 64.7 & 74.2 & 64.2 & 74.1 & 49.2 & 70.2 \\
OCRB & 489 & 597 & 489 & \cellcolor{gray!12}590 & 394 & 512 & 396 & 523 & 264 & 426 \\
\midrule
Ret. & 88.8 & 92.5 & 89.4 & \cellcolor{gray!12}93.3 & 85.3 & 89.1 & 85.0 & 89.9 & 75.8 & 83.8 \\[-2pt]
 & $\pm$0.2 & $\pm$0.4 & $\pm$0.3 & \cellcolor{gray!12}$\pm$0.4 & $\pm$0.2 & $\pm$0.6 & $\pm$0.3 & $\pm$0.2 &  &  \\
TFLOPs & 373 & 373 & 65 & \cellcolor{gray!12}65 & 373 & 373 & 65 & 65 & 373 & 373 \\
\bottomrule
\end{tabular}%
}
\end{table}

\addtocounter{table}{1}

\paragraph{Implementation.} We evaluate with VLMEvalKit~\cite{duan2024vlmevalkit} and
lmms-eval~\cite{zhang2025lmmseval} using greedy decoding. MMBench groups that rule matching cannot resolve are
adjudicated by fixed, judge-free rules (\cref{sec:supp-proto}).

\subsection{Main Results}
\label{sec:main}

\paragraph{Frozen MLLM.} With the MLLM untouched, ETD keeps 82.0\% at strict 5\%, 10.7 points above DivPrune, the
strongest training-free baseline (\cref{tab:main}). The learned modules of LLaVA-Mini, LearnPruner and
MetaCompress stay below DivPrune. Importance-based methods land between 52.9 and 67.0\% and
lose most of the text (TextVQA 17 to 27). Downscaling each image to the budget retains 72.7\%; a native-resolution crop of the scorer's cells adds only 0.3 (\cref{sec:supp-bl}). ETD leads at
every budget from 1\% to 25\%, widest at 5\% (\cref{sec:sweep}).

\paragraph{MLLM adaptation.} Trained rows are means over three seeds (\cref{tab:seeds}). LT-OPD w/o merge retains 74.3\%, and 80.5\% as released with its merge; inserting ETD into the
released checkpoint without retraining raises it to 87.8\% (\cref{sec:supp-ltopd}). Our training recipe
retains 85.6\% with DivPrune and 90.5\% with ETD, and full Glance-ETD 91.6$\pm$0.3\%, 6.0 points above the recipe with DivPrune. EPIC retains 63.0\%, below DivPrune, mostly on reading.

\paragraph{Other MLLMs.} On Gemma~4 E4B-it~\cite{gemma4} with the MLLM frozen, ETD keeps 78.6\%, 11.8 points
above DivPrune and above every other training-free method (45.1 to 60.7\%). With MLLM adaptation
Glance-ETD retains 83.6\% and the dense-encoder variant 84.2\%, against 68.7\% for LT-OPD w/o merge (\cref{tab:main}).

\subsection{Ablation Study}
\label{sec:components}

\begin{figure*}[t]
\centering
\includegraphics[width=\textwidth]{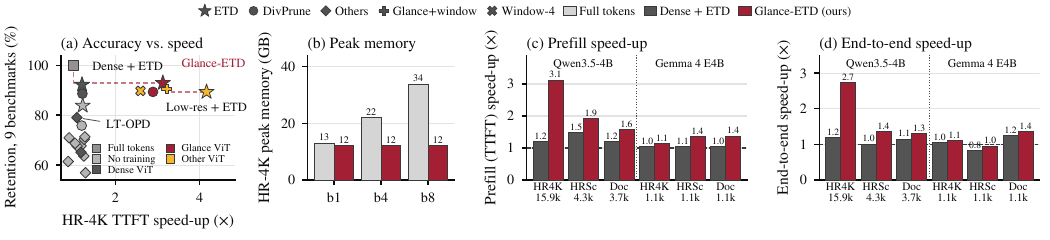}
\caption{Efficiency at strict 5\%, measured on one A800 in bf16. (a) Retention over the nine benchmarks against
TTFT speed-up on HR-Bench 4K; the dashed line is the Pareto front. (b) Peak memory on HR-Bench 4K at batch 1, 4
and 8. (c) TTFT and (d) end-to-end speed-up at batch 1, on three benchmarks per model ordered by image size.
\Cref{tab:latency-native} covers every benchmark of both models, and \cref{tab:latency,fig:eff-supp} the rest of
the measurements.}
\label{fig:efficiency}
\end{figure*}

\Cref{tab:components} removes each stage at strict 5\%. The 10$\times$ encoder learning rate alone adds 3.5 points to DivPrune (85.3 to 88.8\%). On
top of that recipe ETD adds 3.7 points to 92.5\%, mostly on TextVQA and OCRBench, which rise from 70.4 to 76.3
and 489 to 597. The glance encoder adds a further 0.8 to 93.3\% and cuts HR-Bench 4K prefill from 373 to 65 TFLOPs and
TTFT from 3355 to 1290\,ms. With the original recipe ETD adds 3.8 points on
the dense encoder and 4.9 on the glance one, the encoders stay within one point of each other (85.3 \vs 85.0 with DivPrune, 89.1 \vs 89.9 with ETD), and the 10$\times$ rate adds 3.4 to Glance-ETD. Restricting the
dense model's candidates to the glance cells leaves it at 92.2\%: the glance gain comes from native
re-encoding (\cref{sec:supp-vit}).

\subsection{Analysis of the Selection Rule}
\label{sec:ablation}

\Cref{tab:ablation} changes one inference choice of Glance-ETD at a time on the seed-0 models, which both retain 92.9\%; drops are on Qwen first and Gemma second. \Cref{sec:supp-abl} describes
every row and repeats seven over three seeds, where the ranking holds.

\paragraph{The glance should follow the question.} Choosing these cells by diversity, the rule used in training,
costs 1.0 and 2.1 points, and random cells cost 2.0 and 2.0. The loss falls on V$^*$ and HR-Bench 4K, where the
answer is a small object: diversity cells take V$^*$ from 82.7 to 77.5.

\paragraph{ETD needs both evidence and diversity.} At $\gamma=0$ ETD is DivPrune and loses 2.9 and 3.2 points,
mostly on reading, where TextVQA falls from 76.8 to 70.8 and OCRBench from 583 to 502. At $\gamma=64$ it keeps
the tokens with the highest $p$ and loses 3.6 and 2.5: they cover the same words several times and miss
the rest of the scene, and GQA falls from 53.8 to 51.7.

\paragraph{How the two are combined.} Splitting the budget in half between evidence and diversity costs 1.6 and 2.6
points, and switching by question type 2.1 and 1.8. Putting the same evidence in the
DPP~\cite{kulesza2012dpp} of CDPruner~\cite{zhang2025cdpruner} costs 1.3 and 1.4, ties ETD only with a tuned
temperature and is four times slower. A mild tilt is enough: every $\gamma$ from 0.125 to 1 stays within one
point of 0.25 on Qwen and two on Gemma, so one $\gamma$ serves both models (\cref{sec:supp-abl}).

\paragraph{The evidence comes from box supervision.} With the MLLM frozen, a scorer without the question
input keeps 5.5 of ETD's 8.0 points over DivPrune, one without boxes none (\cref{tab:supervision}). Giving the scorer another question of the same benchmark costs 2.2 and 2.1.
With no TextVQA, TextCaps, DocVQA, InfoVQA or GQA data in its training set it still gives 93.0\% (\cref{sec:supp-dedup}). On TextVQA, ETD keeps a token inside the answer box for 99.1\% of questions against 80.2\% for DivPrune (\cref{fig:motivation}c, \cref{sec:diag}).

\subsection{Efficiency}
\label{sec:latency}

\Cref{fig:efficiency} reports speed and memory (\cref{sec:supp-acc}). On HR-Bench 4K the ViT is 2.64\,s of a
4.03\,s TTFT, so selectors after it stay at 0.87 to 1.28$\times$ (\cref{tab:latency-bl}). Glance re-encodes only
the scorer-chosen cells, cutting the ViT to 0.57\,s and prefill from 482 to 65 TFLOPs: TTFT is 3.13$\times$ and
end-to-end time 2.73$\times$ shorter, batch-8 throughput 2.53$\times$ higher, and memory 12.1 instead of
33.9\,GB (\cref{tab:latency}). The gain grows with image size (\cref{tab:latency-native}): TTFT is 2.65 to
3.13$\times$ shorter on HR-Bench and 1.38 to 1.92$\times$ on V$^*$, DocVQA, HRScene and MME-RealWorld, but 0.85
to 0.95$\times$ below the glance threshold; on Gemma, which resizes images to about 1.1k tokens, it is
1.13 to 1.51$\times$.

\begin{figure}[t]
\centering
\includegraphics[width=\linewidth]{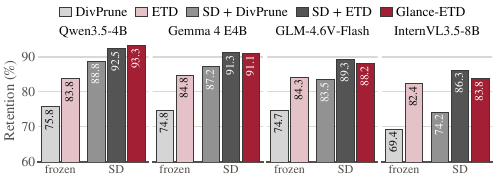}
\caption{Retention over nine benchmarks at strict 5\% on four MLLMs. SD is self-distillation with the dense
encoder; \cref{tab:xmodel} gives the numbers.}
\label{fig:xbackbone}
\end{figure}

\subsection{Cross-Backbone Generalization}
\label{sec:xbackbone}

\Cref{fig:xbackbone} repeats the comparison on GLM-4.6V-Flash~\cite{hong2025glmv} and
InternVL3.5-8B~\cite{wang2025internvl35}, which splits images into crops, each with its own scorer (\cref{sec:xmodel}). ETD leads DivPrune on every model, by 8.0 to 13.0 points frozen and 3.7 to 12.0 after self-distillation. Glance-ETD stays above self-distillation with DivPrune on all four, within 2.4 points of the dense one.

\section{Conclusion}
\label{sec:conclusion}
One scorer drives the glance encoder and the selection. At 5\% of the tokens, Glance-ETD retains 91.6\% of Qwen3.5-4B over fourteen benchmarks, 17.3 points above the prior best, and reaches the first token 3.1$\times$ sooner on 4K images.

\clearpage
\section*{LLM Usage Statement}
We used Claude Code (Anthropic, Claude models) as a coding and writing assistant. It wrote and ran parts of the
training, evaluation and analysis scripts under the authors' direction, generated the LaTeX tables from the
evaluation outputs with scripts, and drafted and revised the manuscript text, which the authors edited and checked.
It did not generate experimental measurements: every reported number comes from evaluation runs of the models, and
MMBench adjudication is rule-based without a judge model. Citations were checked against the source papers by the
authors. No generative model is an author of this paper.

{
 \small
 \urlstyle{same}
 \bibliographystyle{ieeenat_fullname}
 \nocite{shao2024grpo,sidorov2020textcaps,dao2022flashattention}
 \bibliography{refs}
}

\clearpage
\setcounter{section}{0}
\setcounter{subsection}{0}
\setcounter{subsubsection}{0}
\setcounter{table}{0}
\setcounter{figure}{0}
\setcounter{equation}{0}
\setcounter{algorithm}{0}
\renewcommand{\thesection}{\Alph{section}}
\renewcommand{\thetable}{S\arabic{table}}
\renewcommand{\thefigure}{S\arabic{figure}}
\renewcommand{\theequation}{S\arabic{equation}}
\renewcommand{\thealgorithm}{S\arabic{algorithm}}
\renewcommand{\theHsection}{supp.\Alph{section}}
\renewcommand{\theHsubsection}{supp.\Alph{section}.\arabic{subsection}}
\renewcommand{\theHtable}{supp.\arabic{table}}
\renewcommand{\theHfigure}{supp.\arabic{figure}}
\renewcommand{\theHequation}{supp.\arabic{equation}}
\renewcommand{\theHalgorithm}{supp.\arabic{algorithm}}
\makeatletter
\setlength{\@fptop}{0pt}
\setlength{\@fpsep}{8pt plus 2fil}
\setlength{\@fpbot}{0pt plus 1fil}
\setlength{\@dblfptop}{0pt}
\setlength{\@dblfpsep}{8pt plus 2fil}
\setlength{\@dblfpbot}{0pt plus 1fil}
\makeatother
\renewcommand{\dblfloatpagefraction}{0.9}
\renewcommand{\floatpagefraction}{0.9}
\renewcommand{\bottomfraction}{0.85}
\insupplementtrue
\maketitlesupplementary
\printsupplementcontents
\section{Glance-ETD inference}
\label{sec:supp-alg}
\Cref{alg:glance-etd} is the inference procedure. Stage~1 runs on images with at least 4096 patches; smaller images
are encoded densely. The same scorer $f_\phi$ is called twice, on the coarse grid and on the merged grid. The first
token is the one least similar to the mean token, as in DivPrune, so a uniform evidence map leaves the loop identical
to DivPrune. On a glanced image $X$ has the size of the dense encoder's output: unselected cells contribute their coarse
token at each of their four native slots, and these tokens stay candidates for Stage~2 (budget accounting in
\cref{sec:supp-proto}).
Each later token maximises \cref{eq:etd} with $\gamma=0.25$. Training does not use the scorer: the MLLM is distilled
with diversity cells and DivPrune, and the scorer is trained separately (\cref{sec:training} describes that training in the main text).

\begin{algorithm}
\algcaption{Glance-ETD inference.}
\label{alg:glance-etd}
\begin{algorithmic}
\Require image, question $q$, rate $r{=}0.05$, $\gamma{=}0.25$
\Ensure $K$ visual tokens at their native positions
\If{at least $4096$ patches}
  \State $\bar X \gets$ encode the $2{\times}2$-pooled image
  \State $(\bar r_i,\cdot) \gets f_\phi(\bar X,q)$
  \State $C \gets$ the $25\%$ of the coarse cells with largest $\bar r_i$
  \State re-encode the native patches of $C$
  \State $X \gets$ second-pass on $C$, coarse copies elsewhere
\Else
  \State $X \gets$ dense encoder output
\EndIf
\State $(r_i,\cdot) \gets f_\phi(X,q)$
\State $p \gets \mathrm{softmax}(r)$
\State $N \gets |X|$,\quad $K \gets \lceil rN\rceil$
\State $S \gets \{i\text{ least similar to the mean of }X\}$
\While{$|S|<K$}
  \State $d_i \gets \min_{j\in S}(1-\cos(x_i,x_j))$, $i\notin S$
  \State $i^\star \gets \arg\max_{i\notin S}\, d_i(Np_i)^{\gamma}$
  \State $S \gets S\cup\{i^\star\}$
\EndWhile
\Return $X_S$
\end{algorithmic}
\end{algorithm}

\section{Protocol and Baselines}
\label{sec:supp-setup}
The entries below fix the seeds, the scoring protocol, the baseline ports, and the cost of the shared training recipe.

\subsection{Training Seeds and Variance}
\label{sec:supp-seeds}
\Cref{tab:main} reports, for our method, the mean over three training seeds.
\Cref{tab:seeds} lists all three. Seeds differ only in the training random seed of each run.

\begin{table*}[t]
\centering
\small
\setlength{\tabcolsep}{2.5pt}
\caption{Three training seeds at the strict 5\% budget of \cref{tab:main}. The main table reports the mean over the three seeds; the best seed (Ret.\ All) is in bold.}
\label{tab:seeds}
\resizebox{\textwidth}{!}{%
\begin{tabular}{lccccccccccccccc|c}
\toprule
Method & Seed & GQA & MMMU & MMB & MME & POPE & V$^*$ & HR-4K & HR-8K & MME-RW & HRScene & TextVQA & OCRB & DocVQA & InfoVQA & Ret.\ All \\
\midrule
Self-distillation + DivPrune & 0 & 54.0 & 47.9 & 73.9 & 1975 & 86.3 & 74.9 & 77.0 & 73.5 & 45.9 & 56.2 & 70.5 & 481 & 68.4 & 36.0 & \textbf{85.7} \\
 & 1 & 54.1 & 48.3 & 73.2 & 2033 & 85.3 & 74.9 & 76.6 & 72.8 & 45.2 & 54.9 & 70.0 & 500 & 67.2 & 35.6 & 85.5 \\
 & 2 & 54.4 & 48.7 & 73.5 & 1957 & 86.1 & 75.4 & 77.8 & 73.0 & 45.3 & 54.8 & 70.6 & 486 & 66.0 & 35.6 & 85.4 \\
\midrule
Glance-ETD & 0 & 53.8 & 49.6 & 75.5 & 1955 & 88.2 & 82.7 & 78.9 & 75.0 & 48.0 & 57.0 & 76.8 & 583 & 82.8 & 48.7 & 91.2 \\
 & 1 & 55.0 & 50.1 & 75.2 & 1998 & 88.1 & 84.3 & 78.9 & 76.0 & 50.0 & 55.0 & 75.5 & 603 & 82.4 & 47.5 & \textbf{91.8} \\
 & 2 & 55.3 & 49.1 & 74.9 & 2001 & 88.4 & 83.8 & 77.9 & 76.6 & 48.3 & 58.1 & 76.8 & 584 & 82.5 & 48.0 & 91.7 \\
\midrule
Dense-encoder variant & 0 & 55.1 & 47.9 & 74.7 & 1991 & 88.0 & 79.6 & 78.1 & 74.5 & 45.3 & 55.5 & 75.9 & 585 & 83.1 & 48.5 & 90.2 \\
 & 1 & 55.1 & 49.7 & 75.1 & 2057 & 88.3 & 79.1 & 76.2 & 72.2 & 45.3 & 57.5 & 76.6 & 603 & 83.7 & 50.4 & \textbf{90.9} \\
 & 2 & 54.8 & 48.9 & 74.7 & 1979 & 87.8 & 78.5 & 77.2 & 73.0 & 46.2 & 55.2 & 76.5 & 602 & 83.8 & 49.9 & 90.4 \\
\bottomrule
\end{tabular}
}
\end{table*}

\subsection{Evaluation Protocol and Scores}
\label{sec:supp-proto}
All models are evaluated without thinking: the chat template pre-fills an empty think block and the answer tag. The
LT-OPD paper normalises by thinking-mode full-token scores from the model card (e.g.\ MMMU 77.6), which no
non-thinking row can approach; we normalise by one fixed non-thinking full-token run of our own instead. MMMU uses the official 128-token answer
cap.

\paragraph{Full-token repeats.} Two further full-token runs under the same protocol give V$^*$ 80.6 / 80.6 / 80.6,
HR-Bench 4K 78.0 / 78.0 / 78.0, GQA 60.0 / 60.0 / 60.0, MMMU 51.3 / 51.3 / 51.2, MMBench 79.0 / 79.2 / 79.2, MME
2218 / 2228 / 2226, POPE 89.87 / 89.93 / 89.89, TextVQA 80.49 / 80.56 / 80.56 and OCRBench 848 / 850 / 847 (the
first run is the denominator). Across runs the scores differ by at most 10 on MME, 3 on OCRBench and 0.2 elsewhere.
Normalising Glance-ETD (seed 0) by any one run gives 92.88, 92.77 or 92.84\% retention, and 92.83\% with the mean
of the three, so the choice of denominator moves retention by at most 0.1 points.

\paragraph{Each model's own full tokens.} Training barely changes full-token accuracy: with every visual token, the
LT-OPD-recipe, 10$\times$ dense-encoder and glance models (seed 0) reach 101.0, 99.6 and 100.7\% of the untrained
full-token scores (MMBench $+3$ to $+3.5$, MME $-41$ to $-105$). Normalised by each model's own full-token scores,
DivPrune / ETD retain 84.4 / 88.6\% on the LT-OPD-recipe model (85.4 / 89.6\% with the fixed denominator),
89.0 / 92.5\% (88.6 / 92.2\%) on the dense-encoder model, and 88.6 / 92.2\% (89.3 / 92.9\%) on the glance model. No
entry moves by more than 1.0 point, and the gain of ETD over DivPrune is unchanged.

\paragraph{Retention by benchmark group.} \Cref{tab:groups} splits the retention of \cref{tab:main} by its three
benchmark groups. Glance-ETD retains 94.1\% on the general benchmarks and 99.8\% on the high-resolution ones, where
V$^*$ and HR-Bench are above the full-token model, and 78.1\% on text and documents, where the largest absolute gap
remains (OCRBench 590 of 848, InfoVQA 48.1 of 79.4, DocVQA 82.6 of 94.6). Under the same training, ETD adds 1.5 and
1.4 points to DivPrune on the first two groups and 13.7 on text and documents, and the glance encoder adds 3.6 on the
high-resolution group.

\begin{table}[t]
\centering
\footnotesize
\setlength{\tabcolsep}{4pt}
\caption{Retention (\%) by benchmark group of \cref{tab:main} at strict 5\% on Qwen3.5-4B. General: GQA, MMMU, MMBench, MME, POPE. High-res.: V$^*$, HR-Bench 4K / 8K, MME-RealWorld, HRScene. Text: TextVQA, OCRBench, DocVQA, InfoVQA. Trained rows are three-seed means.}
\label{tab:groups}
\begin{tabular}{@{}lcccc@{}}
\toprule
Method & General & High-res. & Text & All \\
\midrule
DivPrune & 81.7 & 87.1 & 38.4 & 71.3 \\
ETD (ours) & 84.6 & 93.2 & 64.9 & 82.0 \\
LT-OPD & 85.7 & 85.4 & 46.2 & 74.3 \\
\midrule
Self-distillation + DivPrune & 92.5 & 94.8 & 65.3 & 85.6 \\
Self-distillation + ETD (ours) & 94.0 & 96.2 & 79.0 & 90.5 \\
Glance-ETD (ours) & 94.1 & 99.8 & 78.1 & 91.6 \\
\bottomrule
\end{tabular}
\end{table}

\paragraph{MMBench adjudication.} Circular-evaluation groups that rule matching cannot resolve are decided by two
fixed rules without a judge model: a group is correct if every answer names the gold letter unambiguously, and wrong
if some unparsable answer contains no reading that names the gold letter. Groups matching neither rule would need a
judge. No row of \cref{tab:main} needs a judge. The rules decide at most 8 of the 1{,}164 groups in any row (0.7\%
of the MMBench score), and Glance-ETD has at most one such group per seed.

\paragraph{Budget accounting.} \Cref{tab:main} and the original LT-OPD code do not use the same budget. The code sets
$K=\max(32,\lceil 0.05N\rceil)$, a floor of 32 visual tokens on every image, so the keep rate is not strictly 5\% and
is greater than 5\%. On the nine benchmarks that floor raises the total from 5.0\% to 7.1\%. The main table removes
the floor, $K=\lceil 0.05N\rceil$, and every strict row has $\sum K/\sum N=5.0\%$. \Cref{tab:rule} repeats the
selection rules under both. Every selector writes a per-image log of $(N,K)$. On the glance model $N$ counts every slot
of the merged grid, including the four copies of each coarse cell, so $N$ and $K$ equal those of the dense encoder on
the same image and every method keeps the same number of tokens. Stage~2 may keep coarse tokens: the selector receives
all $N$ slots and no quota divides $K$ between second-pass and coarse tokens. The four copies of a coarse cell are
identical, so once one is kept the others have $d_i=0$ and at most one is kept. With 25\% of the cells re-encoded,
a glanced image therefore offers about $0.25N$ distinct second-pass tokens and $0.75N/4\approx0.19N$ distinct coarse
tokens to the $K=0.05N$ budget.

\subsection{Baseline Ports from Code}
\label{sec:supp-bl}
All training-free baselines are ported from the official code and select tokens after the vision merger and before
the LLM. ZOO-Prune perturbs the inputs of Qwen's patch merger, its vision-to-language projector. VisionTrim ranks
tokens by the attention they receive in the penultimate ViT block (Qwen's encoder has no CLS token) and fills the
rest of its budget with the tokens closest to the question. HAWK scores tokens with the pre-RoPE queries and keys of
the first full-attention LLM layer, since the first layers of Qwen3.5 use linear attention; its head weights are
recomputed for this layer, following the procedure given in \cref{sec:supp-hawk}.

The three learned modules are trained on LT-OPD-14K with the MLLM frozen and select before the LLM, like every other
row. LearnPruner trains its 0.53M pruning module with the next-token loss and its ratio penalty (AdamW, $10^{-3}$, batch
32, two epochs); only its first stage is used, since its second stage prunes inside the LLM and would pass more than
$K$ tokens to the first layers. EPIC fine-tunes the whole model for one epoch (LLM $2\times10^{-5}$, encoder
$2\times10^{-6}$, global batch 128) with its token-consistency distillation and a compression ratio annealed to 5\%;
its DART compressor is moved in front of the LLM so that the strict budget counts the same tokens.

\paragraph{Downscaling without pruning.} The simplest way to meet a token budget is to shrink the image until it
yields the budgeted number of tokens and keep all of them, which also lowers the encoder cost. Resizing every image
for the untrained model to 5\% of its visual tokens ($\sum K/\sum N=4.99\%$) retains 74.9\% on the nine LT-OPD
benchmarks and 72.7\% on all fourteen, against 75.8 / 71.3\% for DivPrune and 83.8 / 82.0\% for ETD. Downscaling
keeps the page layout and is strong on documents (DocVQA 67.8, DivPrune 32.1, ETD 71.9), but it loses small objects
and small text (V$^*$ 54.5 \vs 78.0, TextVQA 42.0 \vs 70.2 for ETD). At 25\% it retains 93.9\% on the nine
benchmarks, between DivPrune (92.0\%) and ETD (95.7\%).

\paragraph{Cropping the evidence.} A crop-and-zoom baseline gives the untrained model, within the same $K$, a
native-resolution crop around the cells that Stage~1 of Glance-ETD picks (the max-sum window of $a\times b$ merged
tokens over the rank-weighted cells) and the whole image downscaled to the remaining tokens; images that do not take
the glance path are only downscaled. It retains 75.2\% on the nine benchmarks (downscaling 74.9\%, ETD 83.8\%),
gaining only on V$^*$ (57.6 \vs 54.5) and HR-Bench 4K (68.5 \vs 67.0), and 61.0\% on the five additional benchmarks
against 68.7\% for downscaling and 78.9\% for ETD. The smaller global view costs most on documents and scenes (DocVQA
52.6 \vs 67.8, HRScene 42.8 \vs 50.1): finding the evidence is not enough when the rest of the image becomes a thumbnail.

\subsection{LT-OPD Baseline Reproduction}
\label{sec:supp-ltopd}
The LT-OPD w/o merge rows of \cref{tab:main}, the LT-OPD curve of \cref{fig:budget} and the LT-OPD row of \cref{tab:floor32} are the LT-OPD-14K recipe with
the learnable token merge turned off, trained on each model. CDPruner still selects the tokens. In the released checkpoint, each discarded token is assigned to
its nearest retained token by cosine similarity, and an MLP adds a bounded residual, initialized at zero, onto that
retained token. We disable this merge, so the language model receives the selected tokens alone (pure index pruning).
Our self-distilled models keep the merge, as the released checkpoint does (\cref{sec:training}); every row passes the
same number of tokens to the language model, but the training-free rows and this one carry no merged residual.
The released Qwen3.5-4B checkpoint keeps the merge. At strict 5\% it retains 85.4\% on
the nine LT-OPD benchmarks (V$^*$ 73.3, HR-Bench 4K 71.5, TextVQA 63.3, OCRBench 419), and at floor-32 91.1\%
(V$^*$ 73.3, HR-Bench 4K 71.6, TextVQA 63.6, OCRBench 572); our reproduction with the merge retains 90.3\% at floor-32. On the five added benchmarks at strict 5\% it scores HR-Bench 8K 67.8, MME-RealWorld 44.4, HRScene 54.7,
DocVQA 48.9 and InfoVQA 27.4; on documents it is below the LT-OPD recipe with DivPrune (DocVQA 52.5) and well
below Glance-ETD on that benchmark (DocVQA score 82.0).

\paragraph{ETD on the LT-OPD checkpoints.} Both LT-OPD checkpoints select with CDPruner. Replacing only the selector
at inference, with no retraining, the released checkpoint retains 85.4\% (CDPruner), 86.2\% (DivPrune) and 89.8\%
(ETD) on the nine benchmarks, and 71.6 / 76.2 / 84.2\% on HR-Bench 8K, MME-RealWorld, HRScene, DocVQA and InfoVQA
(TextVQA 63.3$\to$75.3, OCRBench 419$\to$506, DocVQA 48.9$\to$76.2, InfoVQA 27.4$\to$42.0). Our reproduction
without the merge (LT-OPD w/o merge in \cref{tab:main}) gives 79.0 / 81.1 / 87.7\% on the nine benchmarks and 65.9 / 71.5 / 83.6\%. ETD
therefore improves LT-OPD as trained, without our recipe or the glance encoder.

\begin{table*}[t]
\centering
\footnotesize
\setlength{\tabcolsep}{2.2pt}
\caption{LT-OPD with and without its learnable token merge on Qwen3.5-4B at a strict 5\% budget (CDPruner selection, as trained). The released checkpoint keeps the merge; w/o merge is the LT-OPD-14K recipe retrained with the merge turned off, the LT-OPD row of \cref{tab:main} and \cref{fig:budget}. Ret.\ as in \cref{tab:main}.}
\label{tab:ltopd-variants}
\resizebox{\textwidth}{!}{%
\begin{tabular}{lccccc|ccccc|cccc|c}
\toprule
 & \multicolumn{5}{c|}{General} & \multicolumn{5}{c|}{Fine-grained high resolution} & \multicolumn{4}{c|}{Text and document} & Ret.\,(\%) \\
\cmidrule(lr){2-6}\cmidrule(lr){7-11}\cmidrule(lr){12-15}
Method & GQA & MMMU & MMB & MME & POPE & V$^*$ & HR-4K & HR-8K & MME-RW & HRScene & TextVQA & OCRB & DocVQA & InfoVQA & All \\
\midrule
\qwenhead{LT-OPD variants} \\
Full tokens (100\%) & 60.0 & 51.3 & 79.0 & 2218 & 89.9 & 80.6 & 78.0 & 72.6 & 48.9 & 62.5 & 80.5 & 848 & 94.6 & 79.4 & 100.0 \\
LT-OPD~\cite{li2026ltopd}, released (with merge) & 52.8 & 49.8 & 73.5 & 1895 & 85.2 & 73.3 & 71.5 & 67.8 & 44.4 & 54.7 & 63.3 & 419 & 48.9 & 27.4 & 80.5 \\
LT-OPD w/o merge & 47.4 & 48.6 & 66.4 & 1771 & 81.9 & 67.0 & 69.6 & 65.1 & 40.5 & 51.4 & 55.8 & 341 & 40.6 & 25.6 & 74.3 \\
\rowcolor{gray!12}\textbf{Glance-ETD (ours)} & 54.7 & 49.6 & 75.2 & 1985 & 88.2 & 83.6 & 78.5 & 75.9 & 48.8 & 56.7 & 76.3 & 590 & 82.6 & 48.1 & \textbf{91.6}$\pm$0.3 \\
\bottomrule
\end{tabular}
}
\end{table*}

\paragraph{Released checkpoint on all fourteen benchmarks.} \Cref{tab:ltopd-variants} lists both LT-OPD variants. Over the fourteen benchmarks the
released checkpoint retains 80.5\% at strict 5\% with its own CDPruner selection and token merge, 82.6\% with DivPrune
and 87.8\% with ETD, against 74.3\% for the row without the merge. Glance-ETD (91.6\%) is therefore 17.3 points above
LT-OPD without its merge, 11.1 above the released LT-OPD model, and 3.8 above the released model with ETD inserted
at inference. Across the budgets of \cref{fig:budget} (5/10/15/25/50\%, nine benchmarks), the released checkpoint retains
85.4/89.7/92.6/94.8/98.3\% and LT-OPD without its merge 79.0/84.1/87.3/91.7/96.5\%.

\paragraph{From LT-OPD to Glance-ETD.} On the nine benchmarks the path from the LT-OPD row to Glance-ETD is
79.0\% (LT-OPD, CDPruner in training and inference) $\to$ 81.1\% (same weights, DivPrune at inference) $\to$ 85.3\%
(trained with DivPrune, one rollout, 256-token cap and the 1.5K spatial samples; \cref{tab:components}, $1\times$ LR)
$\to$ 88.8\% ($10\times$ encoder learning rate) $\to$ 92.5\% (ETD) $\to$ 93.3\% (glance encoder). Of the 14.3 points,
2.1 come from DivPrune at inference, 4.2 from the step to our LT-OPD-recipe models, 3.5 from the encoder learning rate,
3.7 from ETD and 0.8 from the glance encoder. The 4.2 points are not the recipe changes. With the token merge kept on
and DivPrune at inference (seed 0), our LT-OPD reproduction trained with CDPruner retains 85.0\%, and replacing
CDPruner by DivPrune in training (86.0), adding the spatial samples (85.1), using one rollout (86.2) and the 256-token
cap (85.5) keep it within 1.2 points. The gap to 81.1\% separates these merge-trained models from the one retrained without the merge.

\subsection{Self-Distillation Training Cost}
\label{sec:supp-eff}
\begin{table}[t]
\centering\small\setlength{\tabcolsep}{4pt}
\caption{Training cost of the self-distillation recipe (LT-OPD-14K, 175 steps, 8 A800 GPUs; CDPruner selection at
the floor-32 budget of the LT-OPD release). Retention here uses a full-token reference with MMMU 52.4 instead
of 51.3, which lowers every value by about 0.2 points relative to the main tables; the comparison within the table
is unaffected.}
\label{tab:supp-eff}
\begin{tabular}{lcc}
\toprule
Rollouts per prompt, answer cap & Training time & Ret.\,(\%) \\
\midrule
8, 1024 (LT-OPD default) & 17.4\,h & 90.1 \\
2, 1024 & 8.7\,h & 91.1 \\
1, 1024 & 7.3\,h & 90.5 \\
1, 256 (ours) & 4.7\,h & 90.7 \\
1, 128 & 4.3\,h & 90.3 \\
\bottomrule
\end{tabular}
\end{table}

\Cref{tab:supp-eff} varies the number of rollouts per prompt and the answer cap. One rollout with a 256-token cap
trains in 4.7\,h instead of 17.4\,h without a loss in retention, and we use it for all trained models. The amount of
data is not a bottleneck either: training on coverage-selected subsets of LT-OPD-14K with the same recipe keeps
90.0--90.7\% from 700 samples (5\%) up to 7K samples, against 90.4\% with all 14K.
The 1.5K spatial samples that replace LT-OPD-14K rows are multiple-choice counting, depth and left--right or
above--below questions, 750 on synthetic indoor scenes and 750 generated from COCO instance annotations; COCO images
used by GQA, POPE or MME are excluded. The LT-OPD row of \cref{tab:main} is trained on the original LT-OPD-14K.

\section{Other Models and Encoders}
\label{sec:supp-models}
The same selectors are measured with a reinforcement-learned scorer and with the other vision encoders reported in this section.

\subsection{Reinforcement Learning of the Scorer}
\label{sec:supp-rl}
We trained the scorer with GRPO~\cite{shao2024grpo}, using the negative teacher KL of the selected subset as a label-free reward and the
split rule as the policy, together with a budget head. At strict 5\% this gives 88.6\% (88.7--88.8\% with the budget
head), against 87.7\% for the supervised scorer under the same rule and 89.6\% for ETD. Variants with group
Plackett--Luce sampling, token-level credit and expert iteration reached 86.1--87.4\%, and GRPO with the greedy subset as
the baseline of the advantage reached 88.5\%. The best of 16 sampled subsets,
chosen by teacher KL, reaches 96.6\%: the reward is informative, but online RL closes about one point of the
8.9-point gap. Distilling the best-of-16 subsets offline into the scorer gives 87.7\%, and 87.3\% when the scorer
also reads query/key features of the frozen LLM, so offline distillation does not close the gap either.

\subsection{Vision-encoder variants}
\label{sec:supp-vit}
\Cref{tab:vit} compares encoders trained with the same recipe and the same 10$\times$ encoder learning rate, all
with ETD and, for the glance encoders, scorer-chosen cells. Relative to glance with the same auxiliary loss at strict 5\%, encoding everything
at low resolution loses 13.7 points on V$^*$ and 6.3 on HR-Bench, and window attention loses 10.3 on HR-Bench, where
it cuts the global context of 4K images. A second-pass window (glance25w) keeps V$^*$ but loses 8.9 on HR-Bench
(90.5\% \vs 91.9\% over nine benchmarks). Over three seeds the glance encoder of the paper is 0.8 points above the
dense encoder at strict 5\% (93.3$\pm$0.4 \vs 92.5$\pm$0.4), ahead on V$^*$ and HR-Bench 4K and level on MMBench
and MME.
To separate the candidate set from the encoding, we restrict Stage~2 on the dense model to the glance pattern: every
token of the top-25\% scorer cells and one token per other $2\times2$ cell, on images of at least 1{,}024 merged tokens.
In a separate three-seed evaluation this changes the dense model's retention from 92.3 to 92.2\% (five held-out
benchmarks: 61.7 to 61.8), while the glance model reaches 93.0\% (62.4). The gain therefore comes from re-encoding the
chosen cells at native resolution, mostly on V$^*$ ($+4$), HR-Bench 8K ($+3$) and MME-RealWorld ($+2.5$).

The encoder variants are trained with the same recipe as the glance encoder and differ only in the encoder path
for images of at least 4096 patches. \emph{Low resolution} encodes the $2\times2$-pooled image only and copies every
coarse token to its four native positions (glance with no second pass). \emph{Window-4} keeps the dense input but
restricts attention to non-overlapping windows of $8\times8$ merged tokens in all blocks except four global ones
(5, 11, 17, 23), as in Qwen2.5-VL~\cite{bai2025qwen25vl}. \emph{Glance25w} is the glance encoder whose second pass uses the same windowed
attention. These variants use the auxiliary loss described next. On MMMU the three-seed means are 48.9 (glance),
48.8 (dense, 10$\times$ LR) and 48.9 (dense, 1$\times$ LR), so the encoder choice does not change the MMMU result here.

\paragraph{Auxiliary loss.} A variant of the glance encoder, the \emph{cosine-regularised} encoder, adds a cosine
loss (weight 0.1) between its merger output and that of the dense encoder. Without it, Glance-ETD retains 93.3$\pm$0.4\% over three seeds instead
of 92.0$\pm$0.1\%, and DivPrune on the glance model 89.4\% instead of 87.7\%. The difference is on MMBench (75.2 \vs
72.7) and MME (1985 \vs 1933), whose images are below the glance threshold and take the dense path: the loss held the
shared encoder close to the frozen dense features. The paper's glance encoder is trained without it. The encoder
variants of \cref{tab:vit}, the zoom-ratio, threshold and
feature analyses of \cref{sec:supp-glance-cov} use the cosine-regularised encoder; each compares settings on the same weights.

\paragraph{Training with ETD.} We also self-distil the glance model with ETD (constant exponent 0.25) instead of
DivPrune selecting the student's tokens, over three seeds. With Glance-ETD at inference it retains 93.5$\pm$0.9\%,
against 93.3$\pm$0.4\% for the model trained with DivPrune; with DivPrune in Stage~2 (Stage~1 unchanged) it retains 89.0$\pm$0.2\%
against 90.3$\pm$0.4\%, lower on reading (TextVQA 67.6 \vs 71.0, OCRBench 464 \vs 505). ETD adds 3.0--4.5 points at
inference whichever selector trained the model, so we train with DivPrune and keep the MLLM independent of the scorer.

\paragraph{HAWK head weights on Qwen3.5.}
\label{sec:supp-hawk}
The released HAWK head weights belong to Qwen2.5-VL-7B. We recompute them for the first full-attention layer of
Qwen3.5-4B with HAWK's procedure (Eq.~3 of~\cite{zhu2026hawk}): each of the 16 heads in turn is blocked from all
visual tokens on the full-token model, and the drop in gold-answer log-likelihood is measured on 300 questions from
each of five LT-OPD-14K training subsets. The weights range from 0.042 to 0.123; head 13 is the most important in
three of the five subsets. At strict 5\% the recomputed weights and uniform weights give the same retention
(56.7\% both), so the HAWK row of \cref{tab:main} does not depend on this choice.

\begin{table*}[t]
\centering
\small
\setlength{\tabcolsep}{4pt}
\caption{Vision-encoder variants, all trained with the same LT-OPD recipe and 10$\times$ ViT learning rate except the first row. Low-res, window and glance variants except the last row add the auxiliary cosine loss (\cref{sec:supp-vit}). Seed 0, selection at inference only.}
\label{tab:vit}
\begin{tabular}{lccccccccc|c}
\toprule
Method & V$^*$ & HR-4K & GQA & MMMU & MMB & MME & POPE & TextVQA & OCRB & Ret.\,(\%) \\
\midrule
\multicolumn{11}{l}{\textit{DivPrune, strict 5\%}} \\
Dense ViT & 73.8 & 76.2 & 52.4 & 50.1 & 71.0 & 1860 & 85.3 & 64.9 & 384 & 85.4 \\
Dense ViT, 10$\times$ LR & 74.9 & 77.0 & 54.0 & 47.9 & 73.9 & 1975 & 86.3 & 70.5 & 481 & 88.6 \\
Low-res (1/4 patches) & 70.2 & 75.1 & 53.5 & 46.9 & 70.4 & 1942 & 85.3 & 70.5 & 504 & 86.9 \\
Glance 25\%, auxiliary loss & 73.3 & 77.4 & 53.6 & 48.1 & 72.6 & 1883 & 84.6 & 70.3 & 475 & 87.5 \\
Glance 25\% + window & 70.2 & 72.5 & 53.7 & 50.4 & 73.2 & 1955 & 85.7 & 70.3 & 501 & 87.8 \\
Window-4 & 75.9 & 68.3 & 53.6 & 49.4 & 71.1 & 1934 & 86.0 & 69.9 & 480 & 87.1 \\
\textbf{Glance 25\%, no auxiliary loss (ours)} & 75.4 & 77.4 & 53.4 & 50.9 & 73.4 & 1971 & 86.6 & 70.3 & 487 & 89.3 \\
\midrule
\multicolumn{11}{l}{\textit{ETD, strict 5\%}} \\
Dense ViT & 78.5 & 77.5 & 53.2 & 49.3 & 72.4 & 1842 & 87.6 & 74.3 & 510 & 89.6 \\
Dense ViT, 10$\times$ LR & 79.6 & 78.1 & 55.1 & 47.9 & 74.7 & 1991 & 88.0 & 75.9 & 585 & 92.2 \\
Low-res (1/4 patches) & 67.5 & 72.6 & 54.6 & 48.0 & 72.5 & 1977 & 87.6 & 75.4 & 594 & 89.3 \\
Glance 25\%, auxiliary loss & 81.2 & 78.9 & 54.8 & 48.0 & 73.0 & 1943 & 87.6 & 76.1 & 580 & 91.9 \\
Glance 25\% + window & 80.1 & 70.0 & 55.3 & 48.9 & 73.3 & 1980 & 87.0 & 75.0 & 562 & 90.5 \\
Window-4 & 76.4 & 68.6 & 55.1 & 47.8 & 73.2 & 1960 & 87.7 & 75.0 & 574 & 89.7 \\
\textbf{Glance 25\%, no auxiliary loss (ours)} & 82.7 & 78.9 & 53.8 & 49.6 & 75.5 & 1955 & 88.2 & 76.8 & 583 & 92.9 \\
\midrule
\multicolumn{11}{l}{\textit{ETD, $\max(32,\lceil 0.05N\rceil)$}} \\
Dense ViT & 78.5 & 77.5 & 57.3 & 52.1 & 81.3 & 2092 & 89.2 & 74.3 & 682 & 95.9 \\
Dense ViT, 10$\times$ LR & 80.1 & 78.0 & 58.1 & 49.9 & 80.9 & 2152 & 88.8 & 76.0 & 709 & 96.6 \\
Low-res (1/4 patches) & 68.1 & 73.4 & 58.0 & 49.6 & 79.6 & 2162 & 89.1 & 75.6 & 732 & 94.4 \\
Glance 25\%, auxiliary loss & 79.6 & 78.9 & 58.2 & 50.3 & 80.6 & 2138 & 89.1 & 76.1 & 730 & 97.0 \\
Glance 25\% + window & 79.6 & 70.2 & 58.5 & 51.2 & 81.1 & 2140 & 89.2 & 74.9 & 688 & 95.4 \\
Window-4 & 76.4 & 68.8 & 58.1 & 51.1 & 80.2 & 2095 & 89.0 & 75.1 & 709 & 94.6 \\
\textbf{Glance 25\%, no auxiliary loss (ours)} & 82.7 & 78.6 & 57.7 & 51.7 & 81.6 & 2168 & 89.1 & 76.8 & 727 & 97.9 \\
\bottomrule
\end{tabular}
\end{table*}

\section{Selection Behavior}
\label{sec:supp-behavior}
These measurements separate which tokens are kept from how many, on the training images and on the evaluation benchmarks.

\subsection{Teacher-KL Selection Analysis}
\label{sec:supp-kl}

\paragraph{Which tokens.} At a fixed $K$ on 8{,}000 training questions, the total cost $\mathcal{L}$ of
\cref{eq:kl} under ETD is 0.46 of DivPrune's: 0.40 on text questions and 1.03 on general ones, and ETD lowers the KL
on 58\% of the questions. Over constant exponents $\gamma\in\{0.25,0.5,1,2,4,8\}$ the ratio rises monotonically from
0.46 to 0.64, so 0.25 is the best of these exponents on training questions; on 2{,}000 validation questions of the scorer data
with the glance model the KL is flat between 0.25 and 0.5 (1.035 and 1.031). The hard split (top-$K/2$ by $p$, the rest by
diversity) reaches 0.64. Choosing the exponent per question by an oracle gives 0.28, and the oracle exponent grows
with $t$ (Spearman $\rho=0.24$; mean 0.68 and 0.82 in the two lower quartiles of $t$, 1.39 in the two upper ones),
but following that trend with $\gamma=t$ gives 0.50, above the constant. Learned temperature heads on $t$, the
concentration of $p$, $\log N$ and other features did not improve over a constant either on the dense model at the
floor-32 budget (\cref{tab:rule}).

\paragraph{How many tokens.} On training questions the per-image KL follows a power law in the kept fraction,
$\mathcal{L}_m(K)\approx c_m(K/N_m)^{-\alpha}$ with $\alpha=0.89$. Minimising $\sum_m\mathcal{L}_m(K_m)$ under
$\sum_m K_m=B$ equalises the marginal gains (water-filling); with $c_m$ independent of $N_m$ this gives
\begin{equation}
  K_m\propto N_m^{\,\eta},\qquad \eta=\alpha/(\alpha+1)\approx 0.47.
  \label{eq:alloc}
\end{equation} Fitting $\eta$ directly on training
questions gives 0.45, and this single power law matches an 8-bin lookup table fitted on the same data (validation KL
0.85 \vs 0.83). Large images receive a smaller fraction because their KL is already low at 5\%. The confidence $t$
does not help predict the budget; answer length does, and a factor of two for open-ended prompts, also fitted on
training prompts, completes the rule used in \cref{tab:alloc} on this model.

\paragraph{Text head.}
\label{sec:supp-text}
Besides $r_i$, the scorer's MLP outputs a text logit $s_i$, trained jointly by a binary cross-entropy (weight 0.2) on
whether the cell touches an OCR word, on records with OCR boxes. It defines the \emph{text confidence}
$t=\sum_j p_j\,\sigma(s_j)\in[0,1]$, the share of the evidence map on text. Glance-ETD uses neither $s$ nor $t$; they
serve only the gated and $t$-scaled variants of \cref{sec:ablation}, and a scorer trained without the text loss gives
the same retention (\cref{sec:supp-dedup}).

\paragraph{Calibration of the text confidence.} We check the scorer's text head on data it was not trained on:
TextVQA val (4{,}123 questions, all TextOCR word boxes) and the validation split of the scorer data (8{,}685 Visual-CoT
and TextOCR questions, 13 sources). Per token, $\sigma(s_i)$ against ``the cell touches an OCR word'' (3.5M tokens,
base rate 0.18) has an expected calibration error of 0.013 and a Brier score of 0.048; in the 0.3--0.8 bins it is
slightly under-confident (\eg 0.45 predicted, 0.53 observed). Per question, $t$ estimates the share of the evidence
map that lies on OCR cells, and on the 4{,}740 questions with word boxes it matches the measured share with a mean
absolute error of 0.05 (Pearson 0.91). As a classifier of text-evidence questions $t$ reaches an AUROC of 0.995 on the
validation sources (median 0.87--0.96 on the text sources, 0.08--0.13 on GQA, Flickr30k, OpenImages, Visual7W and VSR),
with question-level ECE 0.11: it is bimodal, and the few questions with $t$ between 0.4 and 0.6 are text questions
90\% of the time on the validation sources. $t$ does not depend on how much text the image contains (median 0.85--0.89 from 1--5 to over 50 OCR
words on TextVQA). The scorer is therefore a well-calibrated text detector; the ablations in \cref{sec:ablation} show
that this calibration is not what drives the retention gain.

\begin{table*}[t]
\centering
\small
\setlength{\tabcolsep}{4pt}
\caption{Selection rule ablation on the LT-OPD + DivPrune model, identical $K$ per image. Top: strict 5\%. Bottom: $\max(32,\lceil 0.05N\rceil)$.}
\label{tab:rule}
\begin{tabular}{lccccccccc|c}
\toprule
Method & V$^*$ & HR-4K & GQA & MMMU & MMB & MME & POPE & TextVQA & OCRB & Ret.\,(\%) \\
\midrule
\multicolumn{11}{l}{\textit{Strict 5\%}} \\
DivPrune ($\gamma=0$) & 73.8 & 76.2 & 52.4 & 50.1 & 71.0 & 1860 & 85.3 & 64.9 & 384 & 85.4 \\
Split: top-$K/2$ by $p$ + DivPrune & 78.0 & 75.9 & 52.4 & 50.0 & 70.9 & 1844 & 85.8 & 71.6 & 455 & 87.7 \\
Split, scorer fine-tuned with GRPO & 75.9 & 77.6 & 52.5 & 50.3 & 72.4 & 1887 & 86.1 & 71.7 & 481 & 88.6 \\
Tilt set by the text confidence, $d_i (Np_i)^{t}$ & 78.0 & 77.6 & 53.2 & 48.7 & 71.8 & 1842 & 87.7 & 73.3 & 486 & 88.8 \\
ETD, $d_i (Np_i)^{0.25}$ & 78.5 & 77.5 & 53.2 & 49.3 & 72.4 & 1842 & 87.6 & 74.3 & 510 & \textbf{89.6} \\
\midrule
\multicolumn{11}{l}{\textit{$\max(32,\lceil 0.05N\rceil)$}} \\
DivPrune ($\gamma=0$) & 73.8 & 76.9 & 57.5 & 51.1 & 80.8 & 2024 & 89.5 & 65.4 & 589 & 92.2 \\
Split: top-$K/2$ by $p$ + DivPrune & 78.0 & 75.9 & 57.1 & 51.3 & 80.4 & 2034 & 88.8 & 71.8 & 652 & 94.2 \\
Gate: split only if $t>0.5$ & 74.3 & 76.4 & 57.4 & 51.7 & 80.9 & 2064 & 89.6 & 71.5 & 653 & 94.2 \\
Soft tilt without $t$: $d_i(Np_i)$ & 78.0 & 77.4 & 55.5 & 51.0 & 79.6 & 2072 & 88.8 & 73.3 & 662 & 94.4 \\
Teacher-calibrated temperature head & 78.5 & 77.1 & 57.2 & 51.9 & 81.6 & 2077 & 89.2 & 74.0 & 680 & 95.7 \\
Tilt $d_i(Np_i)^{\beta t}$, $\beta=0.5$ & 77.0 & 77.2 & 57.5 & 51.8 & 80.5 & 2087 & 88.8 & 73.7 & 677 & 95.3 \\
Tilt $d_i(Np_i)^{\beta t}$, $\beta=1$ & 78.5 & 78.1 & 57.7 & 51.8 & 80.8 & 2090 & 89.2 & 73.2 & 666 & 95.5 \\
Tilt $d_i(Np_i)^{\beta t}$, $\beta=2$ & 79.6 & 78.0 & 57.3 & 50.6 & 80.2 & 2082 & 89.3 & 72.8 & 648 & 94.9 \\
ETD, $d_i (Np_i)^{0.25}$ & 78.5 & 77.5 & 57.3 & 52.1 & 81.3 & 2092 & 89.2 & 74.3 & 682 & \textbf{95.9} \\
\bottomrule
\end{tabular}
\\[2pt]\parbox{\textwidth}{\footnotesize Floor-32 rows other than $\beta=1$ and ETD come from a separate run of the same protocol.}
\end{table*}

\subsection{Settings of the selection-rule ablations}
\label{sec:supp-abl}
Every row of \cref{tab:ablation} evaluates the seed-0 model of each backbone with one switch changed at inference.
The MLLM weights, the scorer, the glance threshold and $K$ of every image are those of the full pipeline, and the
Qwen and Gemma inference code implement the same switches. Every max--min rule starts from the token least similar
to the mean token, as DivPrune does in the LT-OPD code.

\paragraph{Three training seeds.} On Qwen we repeated seven rows of \cref{tab:ablation} on the seed-1 and seed-2
models. Retention over the nine benchmarks (mean $\pm$ std, change from the full pipeline): full pipeline
93.3$\pm$0.4; diversity cells 92.0$\pm$0.3 ($-1.3$); random cells 91.1$\pm$0.2 ($-2.2$); DivPrune
90.3$\pm$0.4 ($-3.0$); $\gamma=1$ 92.4$\pm$0.4 ($-0.9$); hard split 91.7$\pm$0.5 ($-1.6$); DPP-$r$
92.5$\pm$0.8 ($-0.8$). The order matches seed 0, and the full pipeline is the best of the seven on every seed.

\paragraph{Stage 1: which cells to re-encode.} For images above the glance threshold the encoder re-encodes 25\% of
the coarse cells at native resolution; the full pipeline takes the cells with the highest relevance $r$, with Stage~2
unchanged. \emph{Max--min diversity} applies \cref{eq:divprune} to the coarse tokens, which is the rule used during
training, so it measures what reading the question at inference adds. \emph{Random cells} draws the same number of
cells uniformly, with a seed fixed per question; it keeps the zoom spread over the image but ignores the question, and
tests whether coverage alone explains the gain. Both rows leave Stage~2 at ETD.

\paragraph{Stage 2: the two factors of ETD.} All Stage-2 rows keep the top-$r$ Stage~1. \emph{W/o evidence} sets
$\gamma=0$, which is DivPrune on the merged native grid. \emph{Strong evidence bias} sets $\gamma=64$; it keeps $d_i$ but lets $p$ dominate. The weights
$(Np_i)^\gamma$ are computed in log space, divided by their maximum and floored at $e^{-80}$ so that they stay positive.
At $\gamma=0.25$ the floor would need $Np_i$ to be $e^{320}$ times below the peak and is never reached; at $\gamma=64$ it
is reached by every token whose $Np_i$ is more than $e^{80/64}\approx3.5$ times below the peak. The row therefore
takes the tokens near the peak of $p$ in order of $p$, and the remaining slots, which share the floor weight, are filled
by distance. It removes diversity among the high-evidence tokens, which is where the full rule spreads the budget over a
text region.

\paragraph{Untilted combinations.} \emph{Hard split} keeps the $\lfloor K/2\rfloor$ tokens with the highest $r$ and
fills the other half by max--min diversity started from them, so the fill avoids tokens close to those already kept.
It is the simplest way to reserve part of the budget for evidence. \emph{Hard gate} uses the split when the text
confidence exceeds 0.5 and plain DivPrune otherwise. Because $t$ is bimodal (\cref{sec:supp-kl}), 0.5 separates
text questions from the rest, so the row asks whether routing by question type helps; ETD has no threshold.

\paragraph{Tilt strength.} The constant exponents 0.125, 0.5 and 1 measure the sensitivity around 0.25. The 0.5 row is
not shown in \cref{tab:ablation}: it gives 92.6\% on Qwen and 91.7\% on Gemma ($-0.3$ and $-1.2$). On the
glance model (seed 0), $\gamma=0.125$, 0.25, 0.5 and 1 retain 76.0, 77.7, 77.2 and 77.3\% at a 1\% budget, 92.4,
92.9, 92.6 and 92.0\% at 5\%, and 99.9, 99.9, 100.0 and 99.0\% at 25\%, so the fixed 0.25 is best at 1\% and 5\% and
within 0.1 of the best at 25\%. Since 0.25 was
chosen by teacher KL on training questions, we also fixed a rule before evaluating these rows: take the $\gamma$
with the lowest mean teacher KL (\cref{eq:kl}) on 2{,}000 validation questions of the scorer data, with the Visual-CoT sources drawn
from GQA, DocVQA and InfographicsVQA left out, using the scorer trained without the text loss. Over
$\gamma\in\{0,0.125,0.25,0.5,1,2\}$ the KL is 1.575, 1.149, 1.035, 1.031, 1.077 and 1.187, so the rule picks 0.5, and
0.5 with that scorer matches the full method over three seeds (\cref{sec:supp-dedup}). \emph{$\gamma=t$} uses the text
confidence of each question as its exponent. It is motivated by the oracle exponent on training questions, which
grows with $t$ (\cref{sec:supp-kl}). The AUROC of 0.90 in \cref{sec:ablation} treats TextVQA and OCRBench questions
as reading and V$^*$, HR-Bench 4K, GQA and POPE questions as perception (29{,}469 questions, glance model).

\paragraph{Source of $p$.} \emph{Wrong question} gives the Stage-2 scorer the question embedding of the preceding
distinct question processed by the same evaluation worker, a question of the same benchmark; Stage~1 and the MLLM
keep the true question. The row tests whether $p$ depends
on the question and not on what is salient in the image. \emph{Question similarity} replaces the scorer by the
z-scored cosine similarity between each visual token and the mean question embedding in the LLM embedding space, the
relevance that CDPruner uses, and keeps the ETD rule with exponent 0.25. It is not a row of \cref{tab:ablation}: it
retains 73.9\% with the MLLM frozen (DivPrune 75.8\%, ETD 83.8\%) and 87.2\% on the glance model with diversity cells
(DivPrune 87.5\%). \emph{DPP-$r$} keeps $p$ and replaces the greedy rule by CDPruner's conditional DPP, a greedy MAP
of $\mathrm{diag}(q)\,L\,\mathrm{diag}(q)$ with $L$ the cosine kernel of the visual tokens and quality $q$ the
min--max normalised $r$. It separates the scorer from the rule. With the quality $q=\exp(\alpha(r-\max r))$ instead,
DPP-$r$ retains 93.0, 92.2, 91.0, 90.5, 89.8 and 88.8\% for $\alpha=0.125,0.25,0.5,1,2,4$ (seed 0): the flattest
quality ties ETD (92.9\%), and over three training seeds the two stay level (93.1$\pm$0.2 against 93.3$\pm$0.4), so
ETD matches a temperature-tuned DPP with a fixed exponent and a simpler greedy rule.
Ranking the tokens by $p$ alone (top-$K$) retains 88.4\%, below the strong evidence bias $\gamma=64$ (89.3\%).

\paragraph{Rows of \cref{tab:rule}.} These rows use the LT-OPD + DivPrune model (dense encoder, LT-OPD recipe) and
use the exponent $\beta t$ instead of a constant. \emph{Tilt set by the text confidence} and \emph{Tilt $d_i(Np_i)^{\beta t}$} use the
exponent $\beta t$. \emph{Teacher-calibrated temperature head} predicts the exponent per image from $t$, the
concentration $1-H(p)/\log N$, the mass of the top $\max(32,\lceil0.05N\rceil)$ tokens of $p$, the fraction of tokens
with positive text logit and $\log N$. It chooses from $\{0,0.25,0.5,1,2,4,8\}$ and is trained to minimise the
expected teacher KL on training questions. \emph{Split, scorer fine-tuned with GRPO} is described in
\cref{sec:supp-rl}.

\subsection{Training-image overlap}
\label{sec:supp-dedup}
The scorer's training images (TextVQA train and Visual-CoT, 167K images) were matched against the images of 17
evaluation sets with 64-bit difference and perceptual hashes (a pair matches when the dHash distance is at most 6
and the pHash distance at most 10). Evaluation images with at least one match (questions on them in parentheses): POPE 132 / 500
(2{,}376 / 9{,}000; COCO images), DocVQA 105 / 1{,}284 (491 / 5{,}349), MMBench 47 / 1{,}115 (187 / 4{,}329), TextVQA
70 / 3{,}166, MME 24 / 1{,}177 (50 / 2{,}374), OCRBench 34 / 930 (35 / 1{,}000) and GQA 1 / 398. The deduplicated scorer is trained identically on data from which every training image matching an
evaluation image is removed (851 images, 1{,}720 of 422K records, mostly Visual-CoT GQA, Visual7W and DocVQA; the
removal rule also adds a pixel-correlation and a 256-bit dHash test).
With the deduplicated scorer in both stages, ETD at strict 5\% retains, on the nine LT-OPD benchmarks, 84.2\% on the untrained MLLM (83.8\% with
the original scorer) and 93.1$\pm$0.7\% with Glance-ETD over three seeds (93.3$\pm$0.4\%). DPP-$r$ with the deduplicated scorer
retains 83.1\% on the untrained MLLM (82.5\%), so ETD stays 1.1 points ahead; with the original scorer the paired
difference is $+1.3$ (95\% interval $[0.4, 2.2]$), mostly from GQA. On the five additional
benchmarks of \cref{tab:main}, HR-Bench 8K, MME-RealWorld, HRScene, DocVQA and InfoVQA, it retains 78.5\% (78.9\%) on the untrained MLLM,
88.6$\pm$0.2\% (88.5$\pm$0.4\%) with Glance-ETD over three seeds, 84.2\% (83.4\%) on the LT-OPD-recipe model and
87.1\% (86.6\%) on the 10$\times$ dense-encoder model; on the trained models no benchmark moves by more than 0.8 except V$^*$ with Glance-ETD (82.0 \vs 83.6, 191 questions). The
LT-OPD-recipe model with ETD scores 76.5 (75.9) on DocVQA, so its gain over DivPrune (52.5) is 24.0, not 23.4.
The overlap does not inflate the reported results on these sets. A stricter scorer is trained without every source that shares a dataset with an
evaluation benchmark (the TextVQA, TextCaps, DocVQA, InfographicsVQA and GQA parts removed; 218K records kept). On the
nine benchmarks it retains 84.0\% (83.8\%) on the untrained MLLM and 93.0$\pm$0.5\% (93.3$\pm$0.4\%) with Glance-ETD
over three seeds. On the five additional benchmarks it retains 78.2\% (78.9\%) and 87.3$\pm$0.6\% (88.5$\pm$0.4\%);
the loss is on DocVQA (82.6$\to$80.1) and InfoVQA (48.1$\to$45.4), whose document training data was removed. A scorer retrained on the features of each trained model, with
the exponent $\beta t$ instead of the constant 0.25, gave 91.5\% on the dense-encoder model (91.9\% with the shared scorer, seed 0) and
$90.0\pm0.4\%$ on the glance model ($90.3\pm0.8\%$), so one shared scorer is used for all Qwen models.
The scorer also needs little box supervision: trained on random 5\%, 15\% or 40\% of the 422K records with
everything else fixed, ETD on the untrained MLLM retains 82.7, 84.2 and 83.9\% (83.8\% with all records; DocVQA 65.7,
70.7 and 72.4 against 71.9), so about 63K records suffice and 5\% loses about one point, mostly on reading.

\paragraph{Same supervision for every selector.} \Cref{tab:supervision} trains the same 1.9M scorer under the
supervision of the learned baselines and runs top-$K$, DPP-$r$ and ETD on each, with the MLLM frozen. With 14K box
records, as many as LT-OPD-14K, ETD retains 81.1\%, two thirds of the gain over DivPrune of the full scorer; without
the question input, which makes it a box-supervised LearnPruner-style scorer, 81.3\%, so 5.5 of the 8.0 points come
from a question-independent evidence prior (text and salient objects) and 2.5 from reading the question. On
Glance-ETD the question-independent scorer, used in both stages, retains 92.3\% over three seeds, against 93.3\% with
the full scorer and 89.4\% with DivPrune, so it keeps 2.9 of the 3.9 points (86.2 \vs 88.5\% on the five additional
benchmarks). Trained on LT-OPD-14K with the
LearnPruner objective and no boxes, the scorer gives no gain under any selector (ETD 75.8\%, equal to DivPrune). The
gain therefore comes from box supervision, not from the size of the scorer or the amount of data. With a box-supervised
scorer, top-$K$ stays 5--8 points below DPP-$r$ and ETD, and ETD is ahead of DPP-$r$ except with 14K records, where
the two are within 0.4.

\begin{table}[t]
\centering
\footnotesize
\setlength{\tabcolsep}{2.6pt}
\caption{Scorer supervision and selector with the MLLM frozen, strict 5\%, retention over the nine LT-OPD benchmarks (DivPrune: 75.8\%). All scorers have the same 1.9M-parameter architecture; 14K = as many box records as LT-OPD-14K; LearnPruner = its objective on LT-OPD-14K.}
\label{tab:supervision}
\begin{tabular}{@{}lccc|ccc@{}}
\toprule
Scorer & Data & Boxes & Question & Top-$K$ & DPP-$r$ & ETD \\
\midrule
Ours & 422K & \checkmark & \checkmark & 77.9 & 82.5 & 83.8 \\
14K box records & 14K & \checkmark & \checkmark & 73.6 & 81.5 & 81.1 \\
No question & 422K & \checkmark & -- & 73.9 & 80.1 & 81.3 \\
No boxes (LearnPruner) & 14K & -- & \checkmark & 57.8 & 74.1 & 75.8 \\
\bottomrule
\end{tabular}
\end{table}

Neither the text head nor in-domain reading supervision is needed. On the glance model with diversity cells (three
seeds), a scorer trained without the text loss, used with $\gamma=0.5$ as chosen in advance by the validation-KL rule
(\cref{sec:ablation}), retains 90.8\%, the same as the full scorer at $\gamma=0.25$ (90.8\%; paired difference
$-0.05$, 95\% interval $[-0.57, 0.45]$) and 3.1 points above DivPrune. Also removing TextVQA and TextCaps~\cite{sidorov2020textcaps} from its
training data changes retention by $+0.3$ on that model and $+0.6$ on the untrained MLLM (seed 0, both intervals
contain zero).

\subsection{Stage-1 cell choice}
\label{sec:supp-glance-cov}
\Cref{tab:s1-profile} changes only how the glance encoder picks its 25\% of coarse cells, at inference, on all three
training seeds; the trained weights, zoom ratio, second pass and token budget are unchanged, so the encoder FLOPs are
the same in every row, and Stage~2 is ETD in every row. \emph{Top-$p$} (ours) takes the cells with the highest scorer
relevance; \emph{diversity} is max--min selection on the coarse tokens, the rule used in training; \emph{query-aware}
runs ETD on the coarse tokens; \emph{half top-$p$} takes half of the cells by relevance and the rest by diversity.
$R_\mathrm{fine}$ is the share of answer-box cells re-encoded at native resolution on the 345 TextVQA questions whose
image takes the glance path. DocVQA and InfoVQA carry no answer boxes, so $R_\mathrm{fine}$ cannot be measured on
them; their accuracy columns stand in for it. A cell the scorer misses is not lost: it keeps its coarse encoding and
remains a Stage-2 candidate, so ETD can still select it. This is why the rules that read the scorer badly are not
catastrophic on documents (random cells reach 80.2 on DocVQA against 82.6 for top-$p$) while the gap is largest where
the evidence is a small object at high resolution (HR-Bench 8K 67.9 \vs 75.9). Over the nine benchmarks 6.1\% of the images (60\% of the visual tokens) take the glance
path: all of V$^*$ and HR-Bench 4K, 25\% of MME, 15\% of OCRBench, 10\% of MMMU, 8\% of TextVQA, and none of GQA,
MMBench or POPE, so the nine-benchmark gain of 1.3 points over diversity comes from few benchmarks (V$^*$ 77.7$\to$83.6, OCRBench
576$\to$590, three-seed means).

Diversity covers text evidence about as well as random cells (32\% \vs 26\%), and on documents it is below random.
Every rule that reads the scorer raises the mean of the four additional benchmarks other than HR-Bench 8K (H4) from 56.5 to 58.2--59.0,
mostly on documents and on MME-RealWorld. The seed-to-seed spread of this mean is at most 0.7. Random
cells raise DocVQA too but are 8 points below top-$p$ on HR-Bench 8K, so coverage alone does not explain the gain. On the
cosine-regularised encoder (\cref{sec:supp-vit}), taking the cells
that touch the answer box first (an oracle, seed 0, with the $\beta t$ exponent in Stage~2) re-encodes 99\% of them and
gains a further 0.4 on glanced TextVQA questions over query-aware cells. The scorer adds about 2\,ms per image, and the
time to first token on HR-Bench 4K is 1.28\,s against 1.39\,s for diversity (full tokens: 4.15\,s). With scorer-chosen
cells the glance model's DocVQA (82.6) is above the dense-encoder model with ETD (80.1, three seeds), so the document
gap of a glance encoder with diversity cells comes from the cell choice, not from encoding a quarter of the image.

\paragraph{Zoom ratio and threshold.} This paragraph and the next use the cosine-regularised encoder. On seed 0, re-encoding 12.5\%, 25\% or 50\% of the cells gives an H4
mean of 55.3 / 55.0 / 57.8 with diversity and 56.0 / 58.2 / 58.9 with query-aware cells, with a time to first token
on HR-Bench 4K of about 1.2--1.3\,s at 12.5\% and 2.0\,s at 50\%. Scorer-chosen cells at 25\% thus match diversity at
50\% for less than three quarters of its encoder time. Applying the glance to every image instead of those above
4096 patches lowers the nine-benchmark retention of the query-aware rule from 91.4\% to 89.6\% (TextVQA 71.3), so
small images stay on the dense path instead of the glance. The threshold is not sensitive: moving it at inference
to 2048, 3072 or 6144 patches (trained at 4096) gives Glance-ETD 91.3, 91.4 and 91.8\% on the nine benchmarks
(91.9\% at 4096, seed 0) and 87.9, 87.8 and 88.3\% on the five additional benchmarks of \cref{tab:main} (88.0\%);
lower thresholds send mid-sized reading images to the glance path (TextVQA 76.1$\to$74.0 at 2048).

\paragraph{Coarse and merged features.} The scorer is trained on dense features but reads the coarse grid in Stage~1
and the merged grid in Stage~2. A Stage-1 scorer trained on $2\times2$-pooled features gives 92.0, 91.9 and 92.4\% over
the three seeds, against 91.9, 92.0 and 92.2\% with the shared scorer. Filling the unselected cells by bilinear
upsampling of the coarse tokens instead of copying each to its four positions gives 91.5\% at inference and 91.6\%
when also used in training (copying: 91.9\%, seed 0). The feature shift does not cost accuracy, and copying is kept.

\begin{table*}[t]
\centering\small\setlength{\tabcolsep}{3.6pt}
\caption{Stage-1 cell rules at inference on the glance model at strict 5\% (mean of three training seeds; Stage~2
is ETD in every row). $R_\mathrm{fine}$: share of answer-box cells of the 345 glanced TextVQA questions that are
re-encoded at native resolution ($^\ddagger$cosine-regularised encoder, seed 0). HR-8K of random: seed 2 only. Ret.: nine LT-OPD benchmarks. H4: mean of DocVQA, InfoVQA, MME-RealWorld
and HRScene. TTFT: HR-Bench 4K, batch 1, one A800 (full tokens 4.15\,s); timed with the cosine-regularised weights, which do not change the compute; half top-$p$ was timed in a second run and scaled by its top-$p$ time.}
\label{tab:s1-profile}
\begin{tabular}{lccccccccc}
\toprule
Stage-1 cells & $R_\mathrm{fine}$ & Ret. & HR-8K & DocVQA & InfoVQA & MME-RW & HRScene & H4 & TTFT (s) \\
\midrule
\rowcolor{gray!12}Top-$p$ (ours) & 88$^\ddagger$ & \textbf{93.3} & \textbf{75.9} & \textbf{82.6} & \textbf{48.1} & \textbf{48.8} & 56.7 & \textbf{59.0} & 1.28 \\
Diversity (as trained) & 32 & 92.0 & 70.3 & 79.5 & 44.8 & 46.0 & 55.6 & 56.5 & 1.39 \\
Random & 26 & 91.1 & 67.9 & 80.2 & 46.3 & 44.8 & 55.4 & 56.7 & 1.31 \\
Query-aware & 73 & 93.0 & 74.9 & 81.4 & 46.6 & 48.0 & \textbf{56.8} & 58.2 & 1.42 \\
Half top-$p$ & 79 & 92.9 & 74.4 & 81.8 & 47.1 & 48.0 & 56.4 & 58.3 & 1.45 \\
\bottomrule
\end{tabular}
\end{table*}

\section{Budgets and Diagnostics}
\label{sec:supp-budgets}
The token budget is varied around the LT-OPD operating point, and the evidence checks include Gemma~4.

\subsection{The LT-OPD operating point}
\label{sec:floor32}
The LT-OPD release sets $K=\max(32,\lceil 0.05N\rceil)$. Small images hit the floor of 32 tokens, so the nine
benchmarks together keep 7.1\% of the visual tokens, not 5.0\%. \Cref{tab:floor32} evaluates each method at
this budget. LT-OPD retains 86.1\%; training with
DivPrune on the same data reaches 92.2\%. Glance-ETD retains $98.0\pm0.1\%$ over three seeds, 11.9 points above
LT-OPD with the same number of tokens, and the dense-encoder variant reaches $96.8\pm0.2\%$.
The released LT-OPD checkpoint, which keeps the learnable token merge, retains 91.1\% (\cref{sec:supp-ltopd}).

\begin{table*}[t]
\centering
\small
\setlength{\tabcolsep}{4pt}
\caption{Operating point of the LT-OPD release, $K=\max(32,\lceil 0.05N\rceil)$ per image ($\sum K/\sum N=7.1\%$). Rows above the line are baselines and LT-OPD-recipe models with DivPrune; rows below are ours.}
\label{tab:floor32}
\begin{tabular}{lccccccccc|c}
\toprule
Method & V$^*$ & HR-4K & GQA & MMMU & MMB & MME & POPE & TextVQA & OCRB & Ret.\,(\%) \\
\midrule
Full tokens & 80.6 & 78.0 & 60.0 & 51.3 & 79.0 & 2218 & 89.9 & 80.5 & 848 & 100.0 \\
DivPrune~\cite{alvar2025divprune}, training-free & 70.7 & 73.9 & 53.4 & 48.8 & 75.5 & 1925 & 86.6 & 49.3 & 431 & 84.1 \\
CDPruner~\cite{zhang2025cdpruner}, training-free & 53.4 & 62.4 & 40.8 & 48.2 & 74.2 & 1862 & 81.9 & 43.5 & 398 & 75.4 \\
LT-OPD~\cite{li2026ltopd} \venue{arXiv'26.09} & 67.0 & 69.6 & 51.7 & 49.4 & 77.8 & 2073 & 85.9 & 56.1 & 533 & 86.1 \\
LT-OPD + DivPrune (ours, reproduced) & 73.8 & 76.9 & 57.5 & 51.1 & 80.8 & 2024 & 89.5 & 65.4 & 589 & 92.2 \\
\midrule
\multicolumn{11}{l}{\textit{Ours and its dense-encoder variants}} \\
\rowcolor{gray!12}\textbf{Glance-ETD (ours)} & 83.8 & 78.4 & 58.1 & 51.7 & 81.4 & 2152 & 89.1 & 76.4 & 727 & \textbf{98.0}$\pm$0.1 \\
\quad ours, dense-encoder variant (no glance) & 79.1 & 77.1 & 58.1 & 50.4 & 80.7 & 2160 & 89.1 & 76.4 & 723 & 96.8$\pm$0.2 \\
\bottomrule
\end{tabular}
\\[2pt]\parbox{\textwidth}{\footnotesize The full-token row is the non-thinking run that every retention in the paper is normalised by.}
\end{table*}

\subsection{Sweep across Token Budgets}
\label{sec:sweep}
\Cref{tab:sweep} varies the strict budget from 1\% to 25\%. On the untrained model ETD is ahead of DivPrune at every
measured budget (+10.3 at 1\%, +8.0 at 5\%, +6.4 at 10\%, +3.7 at 25\%) and ahead of DART by a wider margin.
CDPruner, PruneSID and ZOO-Prune also stay below DivPrune at 10\% (74.3--80.5) and 25\% (84.4--90.3). ETD at
10\% (90.1\%) is within 1.9 points of DivPrune at 25\% (92.0\%) with 2.5$\times$ fewer tokens. After training,
ETD at 3.5\% (86.7\%) is above DivPrune at 5\% (85.4\%). On the glance model, where the ETD row also uses scorer-chosen cells and
the DivPrune row diversity cells, the gap falls from 9.2 points at 1\% to 3.6 at 5\% and 1.7 at 25\%: once a quarter
of the tokens is kept, diversity alone already covers most evidence.

\begin{table}[t]
\centering
\footnotesize
\setlength{\tabcolsep}{2pt}
\caption{Retention (\%) versus strict budget $\sum K/\sum N$.}
\label{tab:sweep}
\begin{tabular}{@{}lcccccccc@{}}
\toprule
 & 1\% & 2\% & 3.5\% & 5\% & 7.5\% & 10\% & 15\% & 25\% \\
\midrule
\multicolumn{9}{@{}l}{\textit{Untrained}} \\
ETD & 66.3 & 75.2 & 81.0 & 83.8 & 87.3 & 90.1 & 93.0 & 95.7 \\
DivPrune & 56.0 & 64.6 & 71.5 & 75.8 & 80.9 & 83.7 & 87.6 & 92.0 \\
DART & 49.2 & 55.5 & 62.4 & 67.7 & 73.2 & 77.2 & 84.4 & 91.0 \\
CDPruner & 51.3 & 58.3 & 63.7 & 67.6 & 71.5 & 74.3 & 78.6 & 84.4 \\
PruneSID & 51.7 & 59.2 & 65.1 & 68.7 & 77.0 & 80.5 & 81.7 & 86.0 \\
ZOO-Prune & 52.9 & 61.2 & 67.2 & 70.9 & 76.2 & 79.3 & 83.4 & 90.3 \\
\midrule
\multicolumn{9}{@{}l}{\textit{LT-OPD}} \\
ETD & 74.2 & 81.6 & 86.7 & 89.6 & 93.1 & 95.1 & 97.1 & 99.4 \\
\midrule
\multicolumn{9}{@{}l}{\textit{Glance}} \\
ETD & 77.7 & 85.3 & 90.0 & 92.9 & 95.6 & 97.1 & 98.4 & 99.9 \\
DivPrune & 68.5 & 77.9 & 85.4 & 89.3 & 92.8 & 94.3 & 96.5 & 98.2 \\
\bottomrule
\end{tabular}
\end{table}

\subsection{Allocating the budget across images}
\label{sec:alloc}

\begin{table}[t]
\centering\small
\setlength{\tabcolsep}{3.5pt}
\caption{Global budget allocation on the LT-OPD model with ETD. The power of $N$ and
the open-prompt factor are fitted on training questions only; the scale is set to match the reference $\sum K/\sum N$ over the benchmarks shown below.}
\label{tab:alloc}
\begin{tabular}{lcc}
\toprule
Allocation & strict 5\% & floor-32 budget (7.1\%) \\
\midrule
Same ratio for every image & 89.6 & 95.9 \\
$K\propto N^{0.45}$ & 94.6 & 96.3 \\
$K\propto N^{0.45}\cdot 2^{[\text{open}]}$ & \textbf{95.1} & \textbf{97.3} \\
\bottomrule
\end{tabular}
\end{table}

\Cref{tab:alloc} applies \cref{eq:alloc} on top of ETD on the dense LT-OPD model. With the same total number of tokens
($\sum K/\sum N$ of 5.00\% and 4.98\%), giving large images a smaller fraction and small images a larger one raises
strict-5\% retention by 5.0 points, and by 5.6 with the open-prompt factor. The gain comes from the small-image
benchmarks (OCRBench 510$\to$732, MMBench 72.4$\to$79.0, MME 1842$\to$2094), while V$^*$, whose images are the
largest, loses 3.6 points. At the floor-32 budget the gain is smaller (+1.4), because the floor already gives small
images more than 5\%. In an offline sweep of $\eta$ composed from the fixed-ratio runs, the training value 0.45 is
also the best evaluation value. On Glance-ETD (seed 0) the same allocation raises strict-5\% retention from 92.9 to
97.0, and to 97.7 with the open-prompt factor, mostly on OCRBench (583$\to$718 and 781), MMBench (75.5$\to$80.4) and
GQA (53.8$\to$57.7); on training-free
ETD with Gemma~4 it is neutral (85.1 \vs 85.2).

\subsection{Evidence diagnostics}
\label{sec:diag}
\Cref{tab:diag} tests whether ETD's gain comes from the evidence the question asks for. Scoring tokens with a
different question lowers TextVQA from 74.3 to 68.9 and OCRBench from 510 to 441 on the LT-OPD model, and from 76.8 to
72.7 and from 583 to 523 on Glance-ETD, so the scorer uses the question. It still stays above DivPrune with the wrong
question (87.5 \vs 85.4\% on the LT-OPD model, 90.7 \vs 90.0\% on Glance-ETD with the same Stage~1), because part of
the evidence map does not depend on the question (\cref{sec:supp-dedup}). Removing the
tokens ETD would have chosen and selecting from the rest is worse than DivPrune (TextVQA 61.6 \vs 64.9 and 69.0 \vs
70.3), so ETD's tokens carry information that the rest of the image does not replace. With 0.1\% of the tokens the
models fall to 55--57\%, close to answering from the question alone with no image, so retention is informative
between about 55 and 100\%.

\begin{table*}[t]
\centering
\small
\setlength{\tabcolsep}{4pt}
\caption{Does ETD keep the evidence? Strict 5\%. \emph{Wrong question}: ETD scores tokens with the question of another sample, the model then answers the true question. \emph{Complement}: ETD-selected tokens are removed from the pool before selecting. \emph{Blind}: 0.1\% of tokens.}
\label{tab:diag}
\begin{tabular}{lccccccccc|c}
\toprule
Method & V$^*$ & HR-4K & GQA & MMMU & MMB & MME & POPE & TextVQA & OCRB & Ret.\,(\%) \\
\midrule
\multicolumn{11}{l}{\textit{LT-OPD + DivPrune model}} \\
DivPrune & 73.8 & 76.2 & 52.4 & 50.1 & 71.0 & 1860 & 85.3 & 64.9 & 384 & 85.4 \\
ETD & 78.5 & 77.5 & 53.2 & 49.3 & 72.4 & 1842 & 87.6 & 74.3 & 510 & 89.6 \\
ETD, wrong question & 77.5 & 78.0 & 51.3 & 49.7 & 71.4 & 1871 & 87.0 & 68.9 & 441 & 87.5 \\
ETD complement & 72.3 & 77.4 & 51.9 & 49.4 & 72.5 & 1907 & 85.7 & 61.6 & 393 & 85.3 \\
Blind (0.1\%) & 51.8 & 59.6 & 35.4 & 46.6 & 46.1 & 1314 & 61.4 & 10.9 & 59 & 55.2 \\
\midrule
\multicolumn{11}{l}{\textit{Glance model (top-$p$ Stage 1 in the ETD rows)}} \\
DivPrune & 75.4 & 77.4 & 53.4 & 50.9 & 73.4 & 1971 & 86.6 & 70.3 & 487 & 89.3 \\
ETD & 82.7 & 78.9 & 53.8 & 49.6 & 75.5 & 1955 & 88.2 & 76.8 & 583 & 92.9 \\
ETD, wrong question & 80.6 & 78.8 & 52.5 & 49.1 & 74.5 & 1956 & 87.7 & 72.7 & 523 & 90.7 \\
ETD complement & 74.3 & 79.1 & 52.9 & 49.9 & 73.6 & 1896 & 86.8 & 69.0 & 490 & 88.6 \\
Blind (0.1\%) & 49.7 & 62.2 & 35.1 & 45.8 & 51.1 & 1333 & 63.7 & 13.3 & 68 & 56.6 \\
\bottomrule
\end{tabular}
\end{table*}

\subsection{Gemma 4 across budgets}
\label{sec:gemma}

\begin{figure}[t]
\centering
\includegraphics[width=\linewidth]{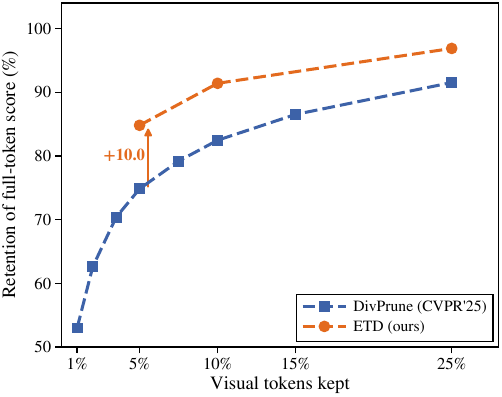}
\caption{Retention of Gemma~4 E4B-it against its own full-token scores. ETD and DivPrune pass the same number of
visual tokens per image. Every marker averages all nine benchmarks.}
\label{fig:gemma}
\end{figure}

\Cref{fig:gemma} extends the Gemma comparison past strict 5\%. ETD retains 91.4\% at 10\%, against 82.5\% for
DivPrune, and 96.9\% at 25\% (DivPrune 91.5\%); ETD at 10\% matches DivPrune at 25\%. At 5\%, TextVQA rises from 38.9 to 59.9 and OCRBench from 184 to 396, against a
full-token OCRBench of 764, and HR-Bench 4K rises from 53.0 to 62.0. GQA, MMMU and POPE stay within 1.0 point of
DivPrune. On Gemma, Glance-ETD uses the two-pass glance with Gemma's own vision tower: a quarter-resolution pass gives
coarse features for every cell, the Gemma scorer picks the 25\% of the cells with the highest relevance, and only
those are re-encoded at native resolution before ETD selects $K$ tokens. It runs on the self-distilled checkpoints,
which are trained without the glance, so unlike Qwen the encoder is not adapted to the glance features. Over three
seeds it retains 91.1$\pm$2.0\% at strict 5\%, against 91.3$\pm$1.1\% for ETD with the dense encoder and
$87.2\pm1.2\%$ for DivPrune; with diversity cells instead of scorer-chosen ones it retains $89.7\pm1.2\%$.

\subsection{Other MLLMs}
\label{sec:xmodel}

\begin{table}[t]
\centering
\footnotesize
\setlength{\tabcolsep}{4pt}
\caption{Retention over the nine LT-OPD benchmarks at strict 5\% on four MLLMs (\cref{fig:xbackbone} plots GLM and InternVL). Each model is compared with its own full-token scores. Qwen and Gemma trained rows are means of three seeds; GLM and InternVL use one seed.}
\label{tab:xmodel}
\newcommand{\xmhead}[2]{\begin{tabular}[b]{@{}c@{}}#1\\#2\end{tabular}}
\begin{tabular}{@{}lcccc@{}}
\toprule
Method & \xmhead{Qwen3.5}{4B} & \xmhead{Gemma 4}{E4B-it} & \xmhead{GLM-4.6V}{Flash} & \xmhead{InternVL3.5}{8B} \\
\midrule
DivPrune & 75.8 & 74.8 & 74.7 & 69.4 \\
ETD & 83.8 & 84.8 & 84.3 & 82.4 \\
\midrule
\begin{tabular}[b]{@{}l@{}}Self-distillation\\+ DivPrune\end{tabular} & 88.8 & 87.2 & 83.5 & 74.2 \\
\begin{tabular}[b]{@{}l@{}}Self-distillation\\+ ETD\end{tabular} & 92.5 & \textbf{91.3} & \textbf{89.3} & \textbf{86.3} \\
Glance-ETD & \textbf{93.3} & 91.1 & 88.2 & 83.8 \\
\bottomrule
\end{tabular}
\end{table}

\Cref{tab:xmodel} repeats the strict-5\% comparison on GLM-4.6V-Flash and InternVL3.5-8B, next to Qwen3.5-4B
and Gemma~4 E4B-it. Each model has its own scorer, trained on the same box data from the features of that untrained
model, and is self-distilled with the recipe of \cref{sec:training} for 175 steps, once with the dense encoder and
once with the glance encoder. InternVL splits an image into crops: its glance encoder encodes the thumbnail, fills
every crop with the upsampled thumbnail region, and re-encodes at full resolution the 25\% of the crops to which the
scorer gives the highest mean relevance. GLM and InternVL rows use one training seed, GQA is scored with the LLaVA
answer normalisation on every row of these two models, and MMBench groups are adjudicated as in \cref{sec:supp-proto}.

ETD is ahead of DivPrune on every model, by 8.0--13.0 points with the MLLM untrained and by 3.6--12.0 after
self-distillation; the largest gaps are on InternVL, where DivPrune keeps 150 of 800 points on OCRBench and ETD 374.
Self-distillation adds 3.9--8.7 points to ETD. The glance encoder is 0.8 points above the dense one on Qwen and 0.2,
1.1 and 2.4 points below it on Gemma, GLM and InternVL. On GLM the loss is on V$^*$ (67.5 \vs 74.9) with
HR-Bench 4K level (71.4 \vs 71.7); on InternVL it is on V$^*$ and HR-Bench 4K (53.4 \vs 64.9, 55.4 \vs 63.0), while POPE
is higher with the glance encoder (72.9 \vs 67.2).

\section{Latency and Memory}
\label{sec:supp-acc}

\begin{figure*}[t]
\centering
\includegraphics[width=\linewidth]{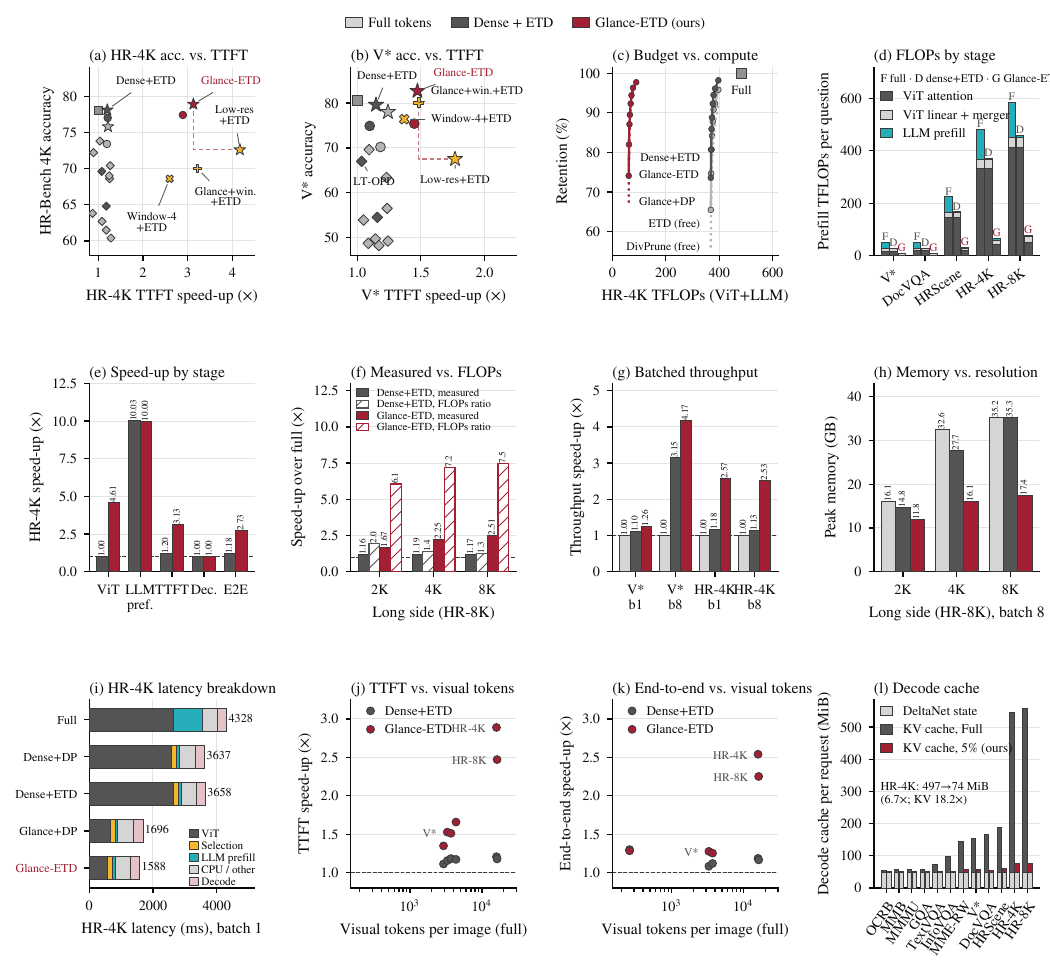}
\caption{Further efficiency views (strict 5\%, Qwen3.5-4B). (a)(b) Task-matched trade-off: HR-Bench 4K and V$^*$
accuracy against TTFT speed-up on the same benchmark (markers as in \cref{fig:efficiency}a). (c) Retention against
HR-Bench 4K prefill TFLOPs as the budget goes from 1\% to 25\% (\cref{tab:sweep}). (d) Prefill FLOPs per question by
stage (F full, D dense + ETD, G Glance-ETD). (e) HR-Bench 4K speed-up of each stage (ViT, LLM prefill, TTFT, decoding,
end to end). (f) Measured TTFT speed-up against the FLOPs ratio (hatched) on HR-Bench 8K resized to 2K--8K.
(g) Throughput speed-up at batch 1 and 8 (\cref{tab:throughput}). (h) Peak memory at batch 8 on resized HR-Bench 8K.
(i) HR-Bench 4K latency by stage at batch 1 for all five configurations (end-to-end ms at the bar ends). (j)(k) TTFT and end-to-end
speed-up against the mean number of visual tokens, on the benchmarks where Glance-ETD exceeds 1.15$\times$. (l) Decode cache per request: KV of the eight
full-attention layers plus the constant Gated-DeltaNet state. (a)--(c) and (e)--(i) use the run of \cref{tab:latency}
for full tokens and the dense and glance encoders; (d) and (j)--(l) use a 4-GPU run that covers all 12 benchmarks,
with speed-ups taken against its own full-token run. The resized sizes in (f)(h) and the V$^*$ bars in (g) were
measured only in the batched-ViT run described in the protocol below.}
\label{fig:eff-supp}
\end{figure*}

\paragraph{Accuracy against cost.} Matched to the benchmark that is timed (\cref{fig:eff-supp}a,b), Glance-ETD
keeps 78.9 on HR-Bench 4K and 82.7 on V$^*$ (seed 0) at 3.1$\times$ and 1.5$\times$ TTFT, while the low-resolution
and windowed encoders give speed for detail. Along the budget axis (\cref{fig:eff-supp}c) a
dense-encoder model moves by at most 30 of its 370--400 TFLOPs between 1\% and 25\%, because the encoder dominates
the cost; the glance encoder moves the whole curve to 60--90 TFLOPs, and Glance-ETD at 7.5\% (95.6\%) uses a
fifth of the FLOPs of the dense model at 5\%.

\paragraph{Where the speed-up appears.} On the twelve benchmarks the TTFT speed-up grows with the number of
visual tokens, as the encoder takes a larger share of the cost: 1.5$\times$ on V$^*$ and DocVQA and
2.5--2.9$\times$ on HR-Bench (\cref{fig:eff-supp}j). End-to-end speed-ups are smaller because decoding takes the
same 59\,ms per token with or without pruning at batch 1 (\cref{fig:eff-supp}e,k, same decode stage).

\paragraph{FLOPs.} We count all matrix multiplications of one TTFT call, including the attention products, from the
checkpoint shapes and the measured token counts of each question: ViT blocks $2(4d^2+2d d_{\mathrm{ff}})$ per patch
plus $4N^2d$ for global attention ($d{=}1024$, $d_{\mathrm{ff}}{=}4096$, 24 blocks), the merger, and the LLM as two
FLOPs per non-embedding parameter (3.57B) per token plus causal attention in its eight full-attention layers; the
Gated-DeltaNet recurrence and the selector are below 0.5\%. On HR-Bench 4K global attention is 332 of the 366 ViT
TFLOPs (\cref{fig:eff-supp}d). Counting only the linear layers gives 138, 41 and 24 TFLOPs for full tokens, dense +
ETD and Glance-ETD, which hides the encoder's quadratic cost. Measured speed-ups stay well below the FLOPs ratio
(2.51$\times$ against 7.5$\times$ at 8K, \cref{fig:eff-supp}f): the LLM prefill alone is 10.0$\times$ faster, but
the ViT only 4.6$\times$, and host-side preprocessing does not shrink (\cref{fig:eff-supp}e).

\paragraph{Memory.} Weights take 9.1\,GB. At batch 1 the remaining peak is set by ViT activations, so selection after
the encoder does not lower it (selection itself adds about 0.3\,GB on HR-Bench 4K). Per
request, the decode cache is $32$\,KiB of KV per kept token in the eight full-attention layers plus a constant
49\,MiB Gated-DeltaNet state (24 layers, fp32 state assumed); at 5\% the KV part falls from 448 to 25\,MiB on
HR-Bench 4K, so the hybrid architecture limits the cache saving to 6.7$\times$ (\cref{fig:eff-supp}l). Peak memory
at batch 8 grows with the image for full tokens and the dense encoder, to 35\,GB at 8K, and stays below 18\,GB with
the glance encoder (\cref{fig:eff-supp}h).

\begin{table}[t]
\centering\small\setlength{\tabcolsep}{2.2pt}
\caption{Efficiency on one A800-SXM4-80GB (bf16, median over 30 questions, configurations run one after another).
HR-4K unless noted; batch 1 unless noted. Times are milliseconds. TFLOPs are ViT + LLM, attention included.
Speed-ups over full tokens are in parentheses. Peak memory is batch 1 / batch 8. Glance-ETD is the final model
with scorer-chosen cells; the HR-4K and batch rows come from the run that timed both, the V$^*$ row from its
companion run with the same protocol.}
\label{tab:latency}
\begin{tabular}{lccc}
\toprule
 & Full & Dense + ETD & Glance-ETD \\
\midrule
TTFT & 4035 & 3355 (1.20$\times$) & 1290 (\textbf{3.13}$\times$) \\
V$^*$ TTFT & 539 & 470 (1.15$\times$) & 366 (1.47$\times$) \\
End-to-end & 4329 & 3658 (1.18$\times$) & 1588 (2.73$\times$) \\
TFLOPs & 366 + 116 & 366 + 6 & 59 + 6 \\
Batch-8 q/s & 0.276 & 0.312 (1.13$\times$) & 0.698 (2.53$\times$) \\
Mem.\ b1/b8 (GB) & 13.0 / 33.9 & 13.3 / 13.3 & 12.1 / \textbf{12.1} \\
\bottomrule
\end{tabular}
\end{table}

\begin{table}[t]
\centering
\footnotesize
\setlength{\tabcolsep}{2.5pt}
\caption{Prefill (TTFT) and end-to-end (E2E) speed-up over full tokens on every benchmark with a latency
measurement, strict 5\% (batch 1, median over 30 questions, one A800). Qwen3.5-4B: same runs as \cref{tab:latency};
Gemma 4 E4B: trained Glance-ETD model (\cref{sec:gemma}), about 1.1k visual tokens per image, configurations interleaved
per question. \Cref{fig:efficiency}c,d shows, per panel, the three benchmarks with the largest Glance-ETD speed-up of each
model.}
\label{tab:latency-native}
\begin{tabular}{l rrrr rrrr}
\toprule
 & \multicolumn{4}{c}{Qwen3.5-4B} & \multicolumn{4}{c}{Gemma 4 E4B} \\
\cmidrule(lr){2-5}\cmidrule(lr){6-9}
 & \multicolumn{2}{c}{Dense + ETD} & \multicolumn{2}{c}{Glance-ETD} & \multicolumn{2}{c}{ETD} & \multicolumn{2}{c}{Glance-ETD} \\
Benchmark & TTFT & E2E & TTFT & E2E & TTFT & E2E & TTFT & E2E \\
\midrule
HR-4K & 1.20 & 1.18 & 3.13 & 2.73 & 1.05 & 1.04 & 1.13 & 1.11 \\
HR-8K & 1.18 & 1.16 & 2.65 & 2.40 & 1.26 & 1.03 & 1.15 & 0.93 \\
HRScene & 1.47 & 1.01 & 1.92 & 1.36 & 1.05 & 0.82 & 1.35 & 0.96 \\
DocVQA & 1.20 & 1.15 & 1.57 & 1.30 & 1.05 & 1.24 & 1.37 & 1.36 \\
V$^*$ & 1.15 & 1.10 & 1.47 & 1.26 & 1.07 & 1.03 & 1.33 & 1.23 \\
MME-RW & 1.13 & 1.04 & 1.38 & 1.13 & 1.24 & 1.07 & 1.51 & 1.23 \\
InfoVQA & 0.95 & 1.05 & 0.95 & 1.14 & 1.04 & 1.06 & 1.39 & 1.25 \\
OCRBench & 0.92 & 1.24 & 0.90 & 1.10 & 1.06 & 1.41 & 1.40 & 1.12 \\
TextVQA & 0.89 & 0.97 & 0.89 & 0.96 & 1.04 & 1.04 & 1.33 & 1.24 \\
GQA & 0.89 & 1.29 & 0.88 & 1.29 & 0.97 & 0.98 & 1.08 & 1.07 \\
MMBench & 0.88 & 0.95 & 0.88 & 0.95 & 0.97 & 0.98 & 1.09 & 1.08 \\
MMMU & 0.89 & 0.98 & 0.85 & 0.95 & 0.97 & 0.98 & 1.10 & 1.08 \\
\bottomrule
\end{tabular}

\end{table}

\Cref{tab:latency} collects the numbers behind \cref{fig:efficiency} on HR-Bench 4K and V$^*$.
\Cref{tab:latency-bl} times every \cref{tab:main} method on Qwen3.5-4B in a separate run with the same protocol.
All of them prune after the dense encoder, so the ViT bounds their speed-up; methods with a heavy selection step
(CDPruner, PruneSID, ZOO-Prune and LT-OPD, whose DPP takes 0.6\,s at 16k tokens) or extra ViT work (VisionTrim) fall
to 0.87--1.09$\times$ against full tokens.

\begin{table}[t]
\centering\footnotesize\setlength{\tabcolsep}{2.0pt}
\caption{Latency of the \cref{tab:main} methods on Qwen3.5-4B at strict 5\% (one A800, bf16, batch 1, median over 30 questions, configurations run one after another; speed-ups over full tokens of the same run). Gain is the speed-up of TTFT over full tokens. TTFT and Sel.\ (selection) in ms, Mem.\ is HR-Bench 4K peak memory in GB. Sel.\ is the selection time on HR-Bench 4K (our implementations; DivPrune and ETD use a CUDA-graph greedy). VisionTrim recomputes a ViT attention map, which the fused attention kernel does not return (1.5\,s of extra ViT time); ZOO-Prune spends 1.35\,s on 64 zeroth-order probes of the projector. Ret.\ is Ret.\ All of \cref{tab:main} (fourteen benchmarks). The Glance-ETD row was timed with diversity-chosen glance cells; choosing them with the scorer adds one scorer call of about 2\,ms (\cref{sec:supp-acc}).}
\label{tab:latency-bl}
\begin{tabular}{lcrrrcc}
\toprule
 & & \multicolumn{3}{c}{HR-Bench 4K} & V$^*$ & \\
\cmidrule(lr){3-5}\cmidrule(lr){6-6}
Method & Ret. & TTFT & Gain & Sel. & Gain & Mem. \\
\midrule
Full tokens & 100.0 & 4099 & -- & 0 & -- & 13.0 \\
\midrule
DivPrune & 71.3 & 3430 & 1.19$\times$ & 141 & 1.18$\times$ & 13.2 \\
DART & 62.8 & 3469 & 1.18$\times$ & 266 & 1.14$\times$ & 13.1 \\
CDPruner & 61.8 & 3776 & 1.09$\times$ & 592 & 1.05$\times$ & 13.0 \\
PruneSID & 65.2 & 3948 & 1.04$\times$ & 769 & 1.09$\times$ & 13.3 \\
VisionTrim & 56.9 & 4696 & 0.87$\times$ & 12 & 1.09$\times$ & 14.7 \\
HAWK & 52.9 & 3190 & 1.28$\times$ & 14 & 1.24$\times$ & 13.2 \\
ZOO-Prune & 67.0 & 4576 & 0.90$\times$ & 1352 & 0.76$\times$ & 13.2 \\
LLaVA-Mini & 66.8 & 3267 & 1.25$\times$ & 12 & 1.24$\times$ & 13.1 \\
LearnPruner & 58.8 & 3290 & 1.25$\times$ & 39 & 1.17$\times$ & 13.1 \\
MetaCompress & 65.9 & 3228 & 1.27$\times$ & 15 & 1.23$\times$ & 13.1 \\
EPIC & 63.0 & 3471 & 1.18$\times$ & 263 & 1.16$\times$ & 13.1 \\
LT-OPD & 74.3 & 3791 & 1.08$\times$ & 581 & 1.04$\times$ & 13.1 \\
\midrule
ETD (ours) & 82.0 & 3362 & 1.22$\times$ & 148 & 1.24$\times$ & 13.2 \\
Glance-ETD (ours) & 91.6 & 1413 & 2.90$\times$ & 162 & 1.48$\times$ & 12.2 \\
\bottomrule
\end{tabular}
\end{table}

\paragraph{Protocol.} All latency numbers come from one run per image size on a single A800-SXM4-80GB (driver
550.54.15, PyTorch 2.10 with CUDA 12.6, FlashAttention~\cite{dao2022flashattention}, bf16), with the five configurations (full tokens, dense
encoder with DivPrune or ETD, glance encoder with DivPrune or ETD) run one after another on the same otherwise idle
GPU. Each configuration sees the same questions (30 per benchmark at native size, 24 HR-Bench 8K questions per
resized size) after three warm-up questions. TTFT is split by CUDA-synchronised hooks into ViT, scorer, selection,
LLM prefill and the remainder (CPU preprocessing and other host work); FLOPs are counted analytically as described below. A second run of the 4K setting on another GPU reproduces every median within 2\%. Full-token p95 on
HR-Bench 4K is inflated by two questions whose prefill takes about 15\,s, which we attribute to kernel compilation
for unseen sequence lengths; the medians are unaffected. Every reported number for full tokens, the dense encoder
with ETD and Glance-ETD comes from three such runs of the final model with scorer-chosen cells, the configuration of
the paper: thirteen benchmarks at native size, HR-Bench 4K also at batch 1, 4 and 8. Their two HR-Bench 4K medians
agree within 1\%. The scorer call takes about 2\,ms, and choosing the cells with it is no slower than by diversity
(1290 against 1367\,ms). The batched-ViT run, which picks the cells by diversity and encodes a whole batch at
once, supplies only what the repeat runs did not measure: the DivPrune, low-resolution and window encoders, the
resized sizes of \cref{tab:size} and the V$^*$ rows of \cref{tab:throughput}.

\Cref{tab:size} varies the image size. The speed-up grows with resolution, from 1.67$\times$ at 2K to
2.51$\times$ at 8K, as the encoder saving dominates; at 8K CPU preprocessing (865\,ms) becomes the largest remaining
cost. The scorer itself takes
2--3\,ms at every size; most of the selection time is the greedy loop. \Cref{tab:throughput} shows that batching
does not help the full-token model at these sizes (its throughput stays flat or drops), while the pruned models gain
from batching. Batch-8 memory on HR-Bench 4K falls from 33.9\,GB to 12.1\,GB, which the dense encoder with ETD
reaches as well (13.3\,GB): the memory saving comes from the short LLM input, not from the glance encoder, whose
gain is the encoder time.

\begin{table}[t]
\centering\footnotesize\setlength{\tabcolsep}{2.5pt}
\caption{Image size. HR-Bench 8K images resized to a long side of 2K--8K pixels (24 questions, batch 1). Glance-ETD
TTFT is split into ViT, scorer plus selection (Sel.), LLM prefill and CPU; speed-up is full / Glance-ETD at the
median. Memory is peak allocated GPU memory of that configuration on the single A800 used for the timing run. Times are in ms.}
\label{tab:size}
\begin{tabular}{lccc}
\toprule
Long side & 2K & 4K & 8K \\
\midrule
Visual tokens $N$ & 2{,}688 & 10{,}752 & 16{,}224 \\
Full TTFT p50/p95 & 385 / 532 & 2283 / --$^\ddagger$ & 4511 / 4719 \\
Dense + ETD TTFT & 331 & 1913 & 3845 \\
Glance-ETD p50/p95 & 230 / 384 & 1015 / 1801 & 1797 / 1855 \\
\quad ViT/Sel./LLM/CPU & 71/20/60/78 & 402/127/91/392 & 678/161/91/865 \\
Speed-up & 1.67$\times$ & 2.25$\times$ & \textbf{2.51}$\times$ \\
Mem. full/glance (GB) & 10.8 / 10.9 & 12.4 / 12.2 & 13.4 / 12.3 \\
\bottomrule
\end{tabular}
\\[2pt]{\footnotesize $^\ddagger$ 15.9\,s, from the compilation outliers described in the text.}

\vspace{10pt}
\setlength{\tabcolsep}{2.5pt}
\caption{Throughput (questions/s, generation up to the official answer cap or 128 tokens) at batch 1, 4 and 8, and
peak memory at batch 8; same run as \cref{tab:latency}. Speed-ups relative to full tokens in parentheses next to each throughput.
$^\dagger$V$^*$ was only batched in the batched-ViT run, which encodes a whole batch at once and chooses the glance cells
by diversity; its pruned configurations therefore hold more activation memory than the HR-4K rows above.}
\label{tab:throughput}
\begin{tabular}{llccc}
\toprule
 & Batch & Full & Dense + ETD & Glance-ETD \\
\midrule
HR-4K & 1 & 0.256 & 0.302 (1.18$\times$) & 0.657 (2.57$\times$) \\
 & 4 & 0.278 & 0.320 (1.15$\times$) & 0.714 (2.57$\times$) \\
 & 8 & 0.276 & 0.312 (1.13$\times$) & 0.698 (\textbf{2.53}$\times$) \\
 & mem. (GB) & 33.9 & 13.3 & 12.1 \\
\midrule
V$^*$$^\dagger$ & 1 & 1.169 & 1.264 (1.08$\times$) & 1.462 (1.25$\times$) \\
 & 4 & 0.910 & 2.229 (2.45$\times$) & 2.873 (3.16$\times$) \\
 & 8 & 0.826 & 2.602 (3.15$\times$) & 3.444 (\textbf{4.17}$\times$) \\
 & mem. (GB) & 17.5 & 16.0 & 12.5 \\
\bottomrule
\end{tabular}
\end{table}

\end{document}